\documentclass[conference]{IEEEtran}
\IEEEoverridecommandlockouts
\usepackage{cite}
\usepackage{amsmath,amssymb,amsfonts}
\usepackage{enumitem}
\usepackage{graphicx}
\usepackage{textcomp}
\usepackage{xcolor}
\def\BibTeX{{\rm B\kern-.05em{\sc i\kern-.025em b}\kern-.08em
    T\kern-.1667em\lower.7ex\hbox{E}\kern-.125emX}}
    
\usepackage{tabu}  
\usepackage[normalem]{ulem}
\usepackage{ulem}
\usepackage[justification=centering]{caption}
\useunder{\uline}{\ul}{}
\usepackage{mathrsfs}
\usepackage[noend]{algpseudocode}
\usepackage{algorithmicx,algorithm}
\usepackage{booktabs} 
\usepackage{multirow} 
\usepackage{bm}       
\usepackage{algorithm}
\usepackage{algorithmicx}
\usepackage{algpseudocode}
\usepackage{balance}
\usepackage{authblk}
\usepackage{float}

\usepackage{array}
\usepackage{url}
\usepackage{threeparttable}

\usepackage[labelformat=simple]{subcaption}

\begin{document}

\title{DEAL: Decremental Energy-Aware\\ Learning in a Federated System
}

\author[1]{Wenting Zou}
\author[2]{Li Li}
\author[1]{Zichen Xu}
\author[3]{Chengzhong Xu}

\affil[1]{Nanchang University}
\affil[2]{Shenzhen Institutes of Advanced Technology, Chinese Academy of Sciences}
\affil[3]{University of Macau}



\maketitle
\begin{abstract}
Federated learning struggles with their heavy
energy footprints on battery-powered devices. The learning 
process keeps  all devices awake while draining expensive battery power to  train a shared model collaboratively,
yet it may still leak sensitive personal information. 
Traditional energy management techniques in system kernel mode  can force the training device entering low power states, but it may violate the SLO
of the collaborative learning. 
To address the conflict between learning SLO and energy efficiency, we propose DEAL, an energy efficient learning system that saves 
energy and preserves privacy with a decremental learning design. DEAL reduces the energy footprint from two layers: 1) an optimization layer
that selects a subset of  workers with sufficient capacity and maximum rewards. 
2) a specified decremental learning algorithm that actively
provides a  decremental and incremental update functions, which allows kernel to correctly tune the  
local DVFS. 
We prototyped DEAL in containerized services with modern smartphone profiles
and evaluated it with several learning benchmarks with realistic traces.
We observed that DEAL achieves 
75.6\%--82.4\%  less  energy footprint in different datasets, compared to the traditional methods. 
All learning processes are faster than  state-of-the-practice FL frameworks up to 2-4X in model convergence.
\end{abstract}
\section{Introduction}~\label{SEC:INTRO}
Federated learning (FL)~\cite{f1}  is a booming distributed learning technique, supporting large, data-intensive deep neural network models.
The FL framework serves when clients refuse to provide personal information while IT companies still want to profile  models, collectively.
As such, the FL technique  supports AI algorithms in many privacy-concerned applications, including Social Network~\cite{f22}, Smart Assistant~\cite{f23} and Traffic Surveillance~\cite{f24}, etc.
The common feature of these applications is that all smart devices involved only communicate with metadata, e.g., model gradients,  to support one cloud to  retrain a global model,  with/without trust. In this way,  the federated learning is claimed to mitigate the impact
of privacy leaking.  Internet giants, like Google, Facebook, adapt FL in their learning framework, in order to conquer the privacy conscious data market with a \$200 billion net  worth~\cite{f25}.


Federated learning supposes to support co-building models among multiple devices, while preserving privacy in-between.
Among these devices, each holds one replica of versioned model and its local data objects. Usually, one small set of
devices, called \emph{servers}, are powerful that can publish a master model, and compute the alignment between all trained local models. Rest of the devices, called \emph{workers}, subscribe models from \emph{servers}, and repeat the training process in-sync, i.e., \emph{federation}.
As such, raw data objects are not exchanged among parties, while the target model may converge after repeating the training process within a sufficient time.
In theory, the federated learning framework is anticipated to scale, working for millions to billions of devices~\cite{f48}.


In practice, this simple FL framework can be expensive and possibly  leaks privacy, for three major reasons. First,
federated learning implicitly assumes that the to-be trained model is simple enough that all synced \emph{workers} can response
and \emph{servers} can compute the alignment within a given time threshold. However, this collaborative learning behavior can reveal
the correlation between similar users, where this similarity model can be reused to reveal some private information. 
Moreover, this model can be complex \emph{per se}, leading to a possible consequence that the model fails to converge.  
Second, \emph{workers}
may spend a large amount of resources on model training,
which  affect the response time and eventually the service-level objective (SLO). Authors of the original federated learning design~\cite{f26}
claimed some strong assumption that models are only trained while plug-in and with WiFi connection. But this claim prevents training data at a fine-grained granularity, violating
the purpose of adapting distributed learning, and ubiquity of mobile devices. In addition, the necessity of being realtime is  a critical performance metric for today's machine learning systems.
Letting device train local fresh data in realtime spends expensive battery energy. 
When scaling-out, this expensive energy footprint
during the training process would grow exponentially, not to mention millions of devices involved, powered by batteries~\cite{f27}.
Thus, a practical federated learning framework should intelligently adapt to privacy concerns with a consideration on the energy efficiency.


To address the aforementioned challenges, it is required to  analyze and further understand the nature
of federated learning. We first provide a study on understanding the state-of-the-practice FL frameworks on several applications. Based on our studies, there are two important observations: (1) For the model complexity and computational cost, in the history of model training, we could expect
the whole framework initiates a simple model with a na\"ive configuration. However, when scaling-out in FL, not all \emph{workers} are active with
the same degree. As such, the FL framework may allow  \emph{workers} to forget some trained features at any time.
That is, models in each \emph{worker} may differ in their versions, which is equivalent to the case that all devices are online but their
model versions are not all updated-to-date. This allows us to provide a tradeoff when publishing aligned models to \emph{workers} at the beginning of each period. We may select a subset of \emph{workers} without waking all device up, in order to reduce energy footprint while still
achieve a possible reward maximization. The optimization process can speed up the  convergence while reducing the  latency, thus local energy footprints. 
(2) To further elaborate the concern on a privacy leak in a federated learning framework, we not only allow each \emph{worker} to remove its old and sensitive data, but also allow
the model to delete the learnt sensitive features, or as we called this framework ``forgets''. This technique is commonly seen as the \emph{decremental learning} in the community.
This feature works in orthogonal with the local energy management in the system kernel. Thus, we can treat it as a control signal to wake the energy management unit when the computation demand decreases, since the device forgets. In this way, all previous  relentless effort on system energy management techniques, such as dynamic voltage and frequency scaling (DVFS), process migration, and IC thermal shutdown, can be adopted into the federated learning process, in order to save training energy. These two observations and derived approaches are the key to provide an energy efficient federated learning system that forgets.

Based on the above findings, we propose a \underline{D}ecremental \underline{E}nergy-\underline{A}ware \underline{L}earning framework (DEAL) that provides an energy efficient design from decremental learning and energy saving techniques.
DEAL serves a two-layered design to reduce the overall energy footprint from the federated learning. When a federated job is created,
DEAL triggers an optimization process that models every candidate device into a multi-armed bandit (MAB) problem, and solves it  to maximize the objective revenue, i.e., training latency, data volume, and
energy footprint. In this way, the whole learning process can be conducted with a performance guarantee. When the learning 
starts, DEAL develops a local middleware layer that carefully manages the local learning process as incremental and decremental updates, based on specific models. The middleware intelligently tunes
local energy state of mobile devices within the decremental learning process. When a bad decision is made in prediction or decremental update, DEAL can resolve the problem by recovering the model in the corresponding decremental and incremental updating algorithm. 
We have prototyped DEAL  in current  mobile operating systems, supporting multiple learning frameworks.
The prototype is evaluated with machine learning models which are widely adopted in real-time machine learning systems,  including Personalized PageRank and Tikhonov Regularization. The evaluation results show that DEAL can significantly reduce all learning completion time up to 2-4X, compared to  the classic federated learning framework. In all state-of-the-practice baselines,
our design shows a 75.6\%--82.4\%  less energy use in all workloads.

\noindent The contribution of our work is summarized as follows,
\begin{itemize}[leftmargin=*]
 \item We identify the high system resource usage issue and privacy problem of a learning federation for mobile systems.
  \item We propose a two-layer energy efficient learning framework, DEAL, which reduces energy footprint with a  forgetting feature. The framework performance is provided with a worst-case mathematical guarantee.
  \item Our prototype proves  the proficiency and effectiveness of DEAL with real world datasets. Compared to conventional federated learning design, DEAL can save 75.6\%--82.4\%  energy cost
  while all learning processes are faster than the classic federated learning framework up to 2-4 orders of magnitude.
\end{itemize}

The rest of the paper is organized as follows. Section \ref{SEC:RISEN} introduces the background about our key concerns that motivate the design of DEAL. Section \ref{SEC:DES} discusses the system design and implementation. Section \ref{SEC:EVA} and \ref{SEC:relatedwork} present the evaluation and related work, respectively. Then Section \ref{SEC:CON} concludes the paper. 
\section{Risen Awareness of Privacy and Resource \\ Use Concern}~\label{SEC:RISEN}

The federated learning (FL) is resource expensive yet may fail to commit the privacy-preserving task. In this section, we first introduce the background of federated learning. After that, we outline the potential privacy leak from collaborative federated learning, with a realworld example. Last, we summarize the  resource use issues in federated learning.

\noindent{\textbf{Federated Learning.}} Federated learning is designed to  train a shared model collaboratively with the data generated on edge devices while preserving data privacy, in a mobile federation.  A federated learning procedure commonly consists of the following steps: 
\begin{enumerate}[leftmargin=*]
    \item At start, the \emph{server} selects a group of mobile candidates, i.e., \emph{workers} to participate in the training process.
    \item The \emph{workers} subscribe the current model and parameters.
    \item  Each \emph{worker} starts local training  with the subscribed model and local data.
    \item When the local training process is completed, each \emph{worker} sends  local coefficients to  the \emph{server}.
    \item The \emph{server} computes the convergence between received gradients. Until the model converges, the process repeats the first step.
\end{enumerate}
Note that, The whole training time is not critical for each individual \emph{worker}, as the job can be throttled. However, the whole training process may prolong exponentially to converge if some \emph{workers}  are delayed for sync. While in-sync,  all training threads keep the device awake, which drains a lot of energy. Therefore, the training completion time   is critical and makes FL very expensive at scale-out.

\begin{figure}[t]
 	\centering
 	\includegraphics[width=0.48\textwidth]{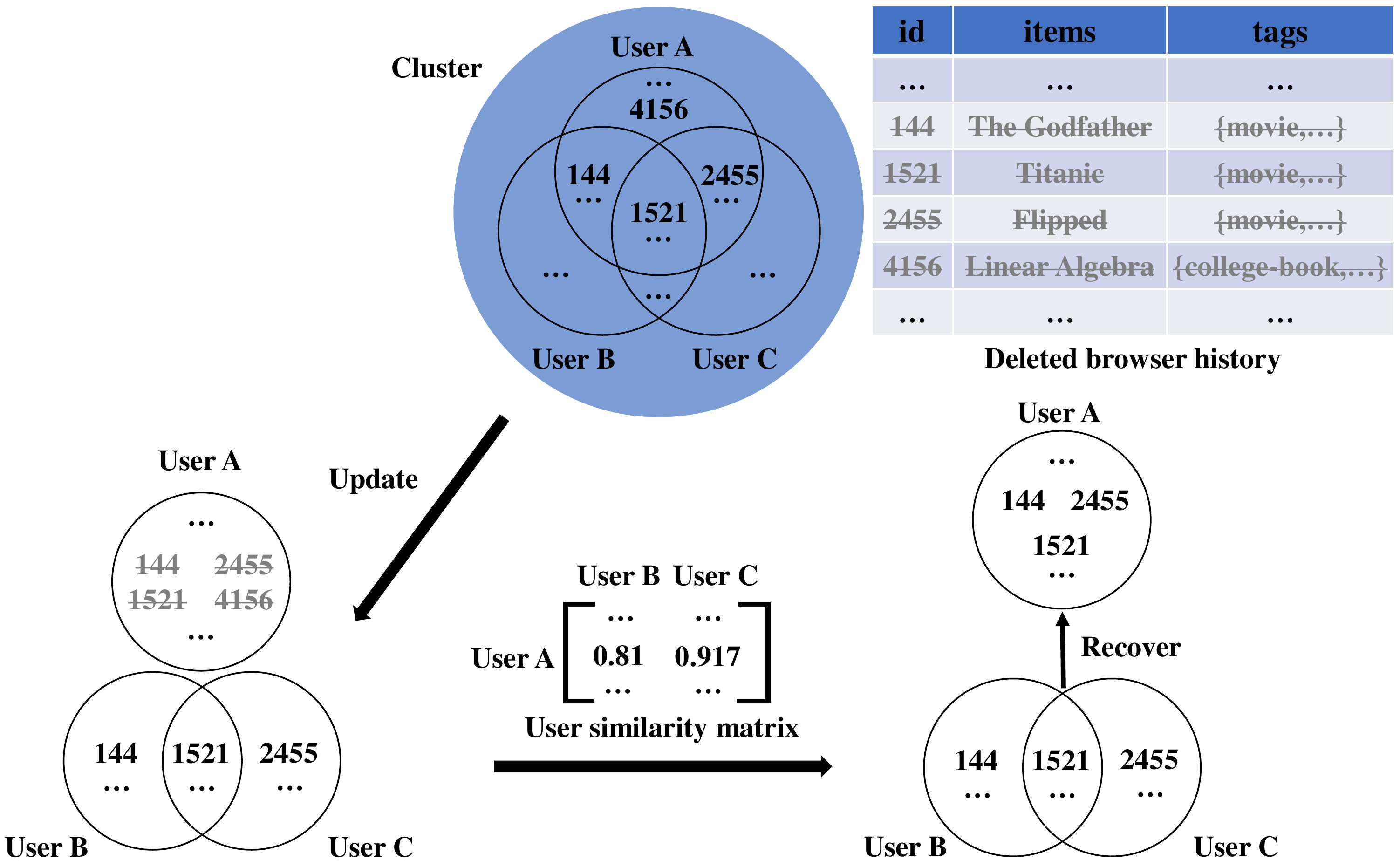}
 	\caption{A possible privacy leak from the Federated Learning, when the original data are deleted. Devices can still reveal the behavior and some privacy guess from user A.}
 	\label{f:moti1}
 	\vspace{-0.2in}
 \end{figure}

\noindent{\textbf{Privacy in Learning.}} Here we   illustrate a real-world example on  privacy leak from the FL process, shown in Figure \ref{f:moti1}. The \emph{Retailrocket}~\cite{f28} e-commerce dataset contains events like  clicks, adding to carts, and transactions over a period of four and a half months, covering  32,000 de-identified  users. Some user (e.g., user A)  has touched the following items: \emph{The Godfather, Titanic, Flipped, and Linear Algebra}. All the information could be potentially sensitive as all items are related to some personal privacy. Regulations, such as European General Data Protection Regulation (GDPR)~\cite{f29}, give users the right   to remove all these sensitive data. However, 
the federated learning, can  still reveal private user information from the database version after removing records from the user A.
For example, FL collects some click and transaction history from all users and computes a similarity matrix between pairs of users, which can be used for personal ranked item recommendation. The similarity distance in this scenario could be computed with a simple Jaccard similarity~\cite{f42}.
Unfortunately, we can still guess this (already deleted)  browsing history from the user A. As shown in Figure \ref{f:moti1}, the similarity matrix shows that the average similarities between user A and some other users are remarkably high, such as the user C  and the user B with a similarity of 0.97 and 0.81, respectively. Digging into this information, we can check the undeleted browsing history from the user B/C to recover the deleted information for user A, on \emph{The Godfather}, \emph{Flipped} and \emph{Titanic}. Therefore, we conclude that it is still possible that these overlapping sets can reveal personal privacy through clustering, which can be considered as a level of privacy leak.



\noindent{\textbf{System Resource Use in  Federated Training.}} 
In the federated training, there are two problems: the idle energy leakage and unnecessary memory footprint. We have already explained in Section~\ref{SEC:INTRO} that all \emph{workers} in the FL framework are usually edge devices, powered by batteries. Previous work \cite{f30} already reveals that a federated learning process consumes heavy energy footprint, shortening the overall device service time by 40\%. Bonawitz et al.~\cite{f9} argue that this is only a technical issue as the learning job is only processed when the device is recharging. However, this assumption violates two important features of the purpose of federated learning: (1) freshness of to-be-trained data objects; (2) the ubiquity of mobile devices. Moreover, that FL only happens when recharging prevents scaling out. Any unnecessary energy leak in a single device may affect all other \emph{workers}, leading to an exponential energy waste. In addition, the learning process needs to repeatedly retrieve all local data from the memory or secondary storage during training, which can cause a large number of page faults, and thus page switches with an extra delay. Thus, in order to have a locally efficiency design, we need to understand the energy footprint and memory use in the local training, and 
an effective middleware that OS can use to  elastically manage the data in memory and allow the learning algorithm coupling with the energy management policy. 

\section{Design and Implementation}~\label{SEC:DES}
Previously, we have discussed our concerns in current FL frameworks on privacy and efficiency. In this section, we   introduce  DEAL, and then analyze our system modeling from the global and local perspectives, and on-device energy control with a decremental learning feature.

\begin{figure}[t]
 	\centering
 	\includegraphics[width=0.48\textwidth]{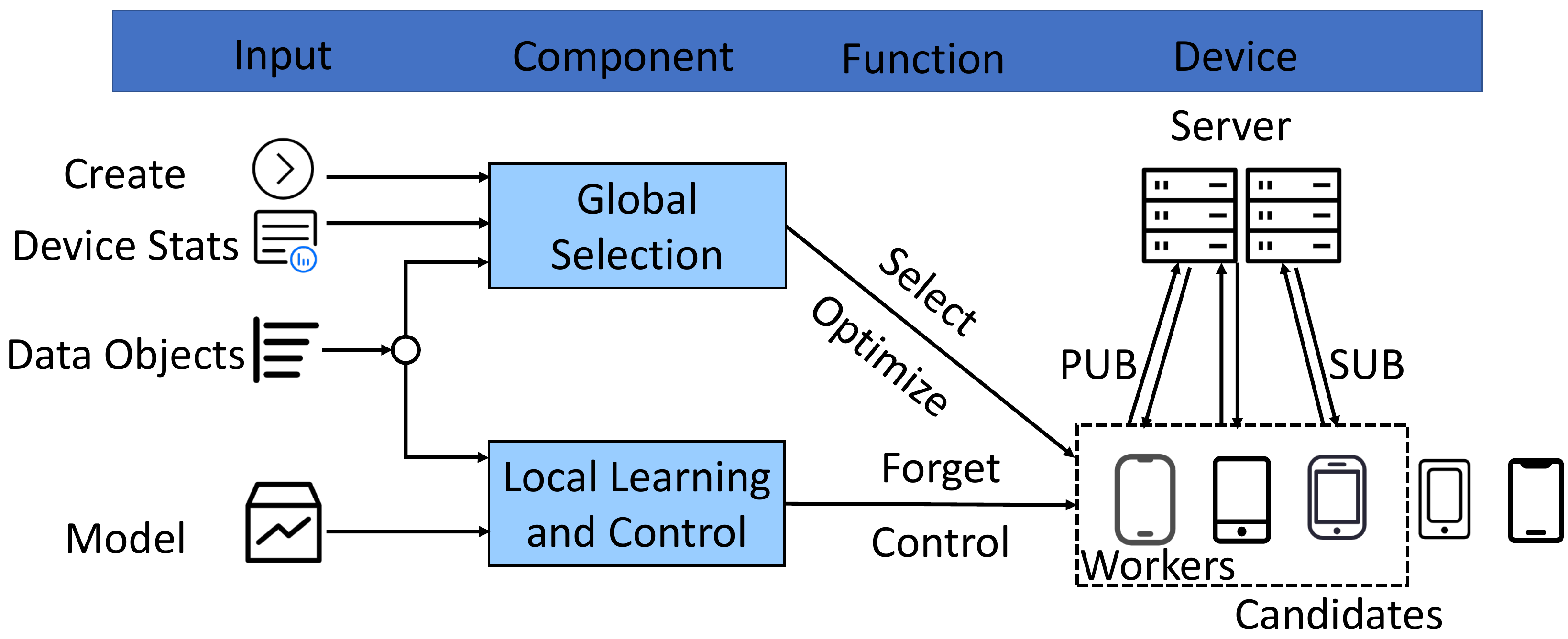}
 	\caption{The system diagram of DEAL.}
 	\label{f:archi}
 	\vspace{-0.2in}
 \end{figure}

\subsection{System Overview}
Figure \ref{f:archi} shows the  architecture of DEAL. DEAL consists of a global layer that provides an  selection optimization process with the MAB algorithm, and a local layer that manages the local decremental learning  through incremental and decremental updates, and the associated energy control. The details are as follows.

\noindent{\textbf{Global Layer.}}
DEAL supports federated learning in a client-server manner. In the global selection layer in the component as shown in Figure  \ref{f:archi}, when a learning job is created from the server, DEAL 
selects a worker subset $\mathcal{N} = \{1,2,...N\}$ from all live candidates. In this subset, all workers shall have required training data $D$, sufficient computation resource to complete the learn job with a reward $X$. The whole selection process shall maximize this reward as an optimization process. As  the Device column in Figure \ref{f:archi}, DEAL initializes the federated learning setup in a PUB/SUB model. All selected workers are notified by the server via the PUB method, as well as receiving the models to be trained. Gradually, each worker finishes its local training, and sends back model gradients via SUB methods. In this process, workers can leave. DEAL allows the server to communicate with workers via the SUB method periodically, and starts the  convergence  process when receiving the majority signals from all selected workers or a Time To Live (TTL) is violated.

\noindent{\textbf{Local layer.}}
In the local layer for  learning and system control, each   worker
introduces a hyperparameter $\theta$, meaning how much one worker shall ``forget'' its data~\cite{f40}. Therefore, though we still compute similarity models between users, as shown in Figure~\ref{f:moti1}, after some epochs, DEAL overwrites the model with newly arrived data and forgets the deleted data, as well as their impact in the model. In this way, we not only allow the balance between model training and local energy reduction, but also enable a better privacy preserved for each worker.

In summary, DEAL exhibits a two-level design, globally and locally,
to optimize the energy efficiency and privacy for federated learning. Next, we introduce our system modeling within these two layers.

\subsection{System Modeling}
As shown in Figure \ref{f:archi}, our global selection process assumes each FL job always starts from the server side. Each server receives SUB signals from all candidate devices, with a profile on the local data volume, available resource, and their
battery capacity. We model all  variables in two categories, namely global and local metrics. In the global metrics, we adapt device statistics, data objects information, etc., in order to find the most suitable subset for one federated learning process. In the local metrics, we describe how we treat local data objects for learning, and models local power footprint, as well as the training time. 

\noindent{\textbf{Global Metrics.}} In the global metrics, DEAL can capture $N$ mobile devices that participate in the federated learning process, as  $\mathcal{N} = \{1,2,...N\}$. Each device $i \in \mathcal{N}$ is related to the reward $X_{i}(k)$ in round $k$, where $k$ = 0,1,2,.... Specifically, the reward is a normalized variable on $[0,1]$ and DEAL can compute the distribution with the the mean represented as $\mu_{i}$. As mentioned above, these reward distributions are \textit{i.i.d}, as defined in a basic FL framework.
The reward vector $\mu = (\mu_{1},..., \mu_{N})$ is unknown before the whole learning process starts.

In each round of learning, devices can join and leave at any given time, due to issues like network outrage, drained batteries, etc. In the PUB/SUB model,
any devices arrived can only subscribe in the next round of learning. Dropped devices are considered as ``sleep'' when violating the TTL $\ddot{T}$ in a learning round.
We use $G(k) \in \mathcal{P(N)}$ to denote the set of available mobile devices in a given round $k$, in which $\mathcal{P(N)}$ is the power set of $\mathcal{N}$. Let $P_{G}(R) \triangleq P(G(k)=R)$, in which $R \in \mathcal{P(N)}$ represents the distribution of the available devices, which is also i.i.d defined in the FL framework. This distribution $R$ is unknown before the training process starts. When
workers start to subscribe, DEAL can reveal $G(k)$ at the server side, in the beginning of each training round $k$.

In each training round, the central server selects $m$ available devices (i.e., the devices belonging to $G(k)$). Each subset of devices creates a worker subset. In order to consider the convergence delay, we ensure that the size of the selected set is not greater than $m$, i.e., the server  starts to compute convergence and update models after selecting $m$ workers to publish. We use ${S(R)}$ to represent the set of all feasible groups when the set of available devices $R$ is observed, i.e., ${S}(R) \triangleq \{ S \subseteq R: |S| \leq m \}$, in which $|S|$ denotes the cardinality of set $S$. In a certain training round $k$, the server not only selects a group $S(k) \in {S}(G(k))$, but also receives a reward represented as $Q(k)$, which is a weighted sum of each participated device's reward, that is, $Q(k) \triangleq \sum_{i\in S(k)}g_{i}X_{i}(k)$, in which $g_{i}$ is the calculated gradient of device $i$. We assume that the gradient $g_{i}$ are fixed positive numbers known, provided from models. The upper bound of $g_{i}$ is represented as $g_{max}>0$. Moreover, DEAL is designed to maximize the expected time-average reward  $E[\frac{1}{K}\sum_{k=0}^{K-1}Q(k)]$ within a given time horizon of $K$ rounds.

\noindent{\textbf{Local Metrics.}} In the local metrics, we first consider the modeling for data objects to be trained.
Let $l = p(D, \theta)$ denote the learnt model, 
where $p$ is a process to learn a model $l$, $D = \left\{d_{1}, d_{2},..., d_{n}\right\}$ denote the training data from $n$ devices, and $\theta$ is the user-defined coefficient. Now, we may need to delete some portion $\theta$ of private data $d_{n}$ of the $n$-th device. The actual learnt model is $l^{'} = p(D \setminus \left\{d_{n}\right\}, \theta)$. This model can be trivially obtained by repeating the training process $p$ on $(D \setminus \left\{d_{n}\right\}, \theta)$, where  $\theta$ defines the percentage of focus on the newly created data objects.

Traditionally, conducting a complete model requires retraining all requested data, which might be costly in energy footprint. Meanwhile, the data loss in a single worker (or a handful of workers, not all in the whole set) may significantly change the predictability of the model. Therefore, we update the existing model $l$ to the desired model $l^{'}$, or update the process $p$ as the process $p_{forget}$. This ``forget'' process $p_{forget}$ can check $d_{n}$ much faster than $p$ for retraining, while still converges, as follows: 
\begin{equation} l^{'} = p_{forget}(p(D, \theta), \left\{d_{n}\right\}, \theta) = p(D \setminus \left\{d_{n}\right\}, \theta) \end{equation}
This methodology is called \emph{decremental learning} ~\cite{f20}, which is similar to online learning, where only the new observations are used to incrementally update the existing model. However, we not only need to use new observations to update incrementally, but also need to delete such updates through the reverse operation.

In DEAL, we consider the practical scenario when deploying FL into real   devices, and mainly focus on the energy and latency from local training and modeling. To understand the training completion time and the estimated energy consumption from one local training, we adopt models from previous research~\cite{f30}. The energy consumption $e$ is a linear combination of device utilization and energy states.
\begin{align}\label{e:power}
  e &=  \int^{T} f_{CPU}\bar{U} + \sum_{j}  e_j
\end{align}
where $\bar{U}$ is the average utilization. $f_{CPU}$ is an energy coefficient with a given frequency. $T$ is the training completion time,
$e_j$ is a static energy profile for each other devices on their specific power states, based on the state-of-the-art state-machine   models~\cite{f39,f45}. 


Next, we model the training completion time. Previous research~\cite{f30} reveals a linear correlation   between local training data size under some model specifications. Hence, the model for training completion time is now simplified as follows:
\begin{equation}
T = A*F(w, \mathcal{M}, D)+B,  \mbox{where}  ~f_{j}~ \mbox{is fixed}
\end{equation}
where, the computation time is positively correlated to a function $F$ of the priority weight $w$, local model $\mathcal{M}$, and affected data size $D_i$, under the current CPU frequency $f_{j}$ in each round of training. $A$ and $B$ are correlation metrics. 
With all aforementioned models, DEAL can provide a decremental learning version of local training based on the data and performance (i.e., energy and latency models). Note that, We need to do corresponding work for each type of specific model. Besides, DEAL may pick wrong devices due to bad prediction. DEAL can only fix the bad selection when another federated learning job initializes. Results show that this may only affect the 95\%-percentile performance. On average, DEAL can sustain a better federated learning service, compared to other state-of-the practice. More discussion on fault tolerance can be found in Section \ref{SEC:EVA}.

\subsection{Global Selection Optimization}
In order to describe the decision of worker selection, we use a binary vector $B(k)=(b_{1}(k),..., b_{N}(k))$ to indicate whether each device is selected to participate in the training in round $k$. $b_{i}(k)=1$ if device $i$ is selected, i.e., $i \in S(k)$; on the other side, $b_{i}(k) = 0$. Then, the action vector $b(k)$ must satisfy $\sum_{i=1}^{N}b_{i}(k) \le m$ for all $k\ge0$.

If a vector of mean reward vector $\mu$ is known in
advance, we can always formulate the reward maximization in federated learning
with minimum selection fraction as an optimization problem~\cite{f46}:

\begin{align}
    \begin{split}
    \mbox{Maximize} &\sum_{R\in P(N)} P_{G}(R) \sum_{S \in S(R)}b_{S}(R) \sum_{i \in S} g_{i}\mu_{i}\\
    \mbox{subject to:} &\sum_{R\in P(N)} P_{G}(R) \sum_{S \in S(R):i\in S} b_{S}(R)  \ge r_{i}, i\in N\\
    &p_i(R) \leq B_i \in [0,1], R \in P(N)
    \end{split}
\end{align}

As the reward vector $\mu$ is unknown, DEAL uses the estimated mean rewards (i.e., exploitation) to maximize the reward and also has to get a more precise estimation of the rewards (i.e., exploration) through simultaneously learning. In this way, we pick an online optimization algorithm to deal with the  exploitation and exploration tradeoff globally.

We treat the DEAL selection optimization  as an Multi-armed Bandit (MAB) problem. MAB problem considers a fixed limited set of resources to be allocated between competing (alternative) choices in a way that maximizes their expected gain, when each choice's properties are only partially known at the time of allocation \cite{f49}. 
In MAB, our main challenges are how to maximize the reward with unknown mean reward distribution and out-of-date worker, and 
how to minimize the power consumption without violating the TTL constraint.
 
For the first critical point to maximize the reward with uncertainty is to fast
retrieve the reward distribution from top workers. We use $c_{i}(k)$ to represent the number of times worker $i$ is selected at the end of round $k$, i.e., $c_{i}(k)\triangleq\sum_{t=0}^{k}b_{i}(t)$. When the system starts at $k=0$, We set $c_{i}(-1)=0$. Meanwhile, let $\widehat{\mu}_{i}(k)$ be the observed mean rewards of worker $i$ by the end of round $k$, i.e., $\widehat{\mu}_{i}(k)\triangleq\frac{\sum^{k}_{t=0}X_{i}(t)b_{i}(t)}{c_{i}(k)}$. If worker $i$ has not been played before the end of round $k$ (i.e., if $c_{i}(k)=0$), we set $\widehat{\mu}_{i}(k)=1$ . We use $\bar{\mu}_{i}(k)$ to represent the estimation of worker $i$ in round $k$, which is given as follows:
\begin{equation}~\label{e:mean}
 \bar{\mu}_{i}(k) \triangleq \min\{\widehat{\mu}_{i}(k-1)+\sqrt{\frac{3\log k}{2c_{i}(k-1)}},1\},
\end{equation}
where $\widehat{\mu}_{i}(k-1)$ and $\sqrt{\frac{3\log k}{2c_{i}(k-1)}}$ correspond to exploitation and exploration, respectively. The upper limit of the above truncated version of the reward estimate is 1, because the actual reward must be [0,1]. Similarly, $\bar{\mu}_{i}(k)=1$ if $c_{i}(k-1)=0$.
Due to the page limit, we refer the detailed optimization analysis in prior work~~\cite{f35}.
Next, we introduce our local decremental learning and energy control to address the local resource problem while forgetting.
 
\subsection{Local Learning and Control}
In order to solve the previously described privacy issues and heavy resource footprint, We propose to use a local decremental learning, which a respect the selected worker set with maximum rewards.
The main idea of our method is that the training algorithm of the target model retains the intermediate results in the process of model calculation. We can update the intermediate results effectively, mainly through two ways: incremental method to merge new user data, and decremental method to delete user data.
As such, we are able to fine tune the local energy configuration, as well as provide a more accurate profile of the local device for next round of global selection. 

DEAL adapts learning algorithms into local training in the following procedure:
\begin{itemize}[leftmargin=*]
    \item \emph{Model Construction}: Construct the prediction model according to the characteristic of the specific learning algorithm.
    \item \emph{Update Procedure}: Design the corresponding decremental and incremental updating algorithms according to the model established in \emph{Model and Prediction}, and derive power savings from it via the dynamic voltage and frequency scaling (DVFS) tuning function.  
    \item \emph{Data Recovery}: Analyze how to recover deleted user data from the stale model. 
\end{itemize}
To adapt a specific learning algorithm into the local DEAL middleware, DEAL first rebuild this algorithm  into a decremental version. This process is usually done offline, and DEAL explores these decremental learning algorithm as a local learning algorithm library. The decremental learning algorithms focus on the incremental/decremental updates, associated with local energy control. When being notified with a degree of ``forget'', or the user-defined variable $\theta$, the DEAL middleware adapts a $\theta$-LRU, that only replaces $\theta$-percent of allocated pages recently used. This algothm can significantly reduces the frequency of page replacement, as well as the number of swaps. 
DEAL keeps track of the level of forgetness in the decremental learning algorithms using data recovery policies, in order to prevent aggressive forgetting and the convergence failure. Next, 
we present DEAL on two learning algorithm cases, namely Personalized PageRank and Tikhonov Regularization, which are widely used in the real-time mobile-based machine learning, in order to highlight the design and implementation of DEAL in the local layer, and show that DEAL can be easily adapted to effectively support other algorithms and systems. 

\begin{algorithm}[t]
\vspace{0in}
  \caption{Update procedure for Personalized PageRank.}
  \label{alg:1}
  \begin{algorithmic}[1]
  \Require concurrency matrix $\textbf{C}$, item interaction counts $\textbf{v}$, similarity matrix $\textbf{L}$, history matrix $\textbf{Y}_{u}$
    \Function{update}{$\textbf{Y}, \textbf{C}, \textbf{v}, \textbf{L}$}
      \State $\textbf{v} \gets \textbf{v} + \textbf{Y}_{u}$
      \While {item pair $(i_{1}, i_{2}) \in \textbf{Y}_{u}$}
        \State $C_{i_{1}i_{2}} \gets C_{i_{1}i_{2}}+1$
      \EndWhile
      \While {item $i_{1} \in \textbf{Y}_{u}$}
        \While {item $i_{2} \in \textbf{C}_{i_{1}}$}
          \State $L_{i_{1}i_{2}} \gets C_{i_{1}i_{2}}/(v_{i_{1}}+v_{i_{2}}-C_{i_{1}i_{2}})$
          \State CPU\_Freq(1)~~//Tune up DVFS
        \EndWhile
      \EndWhile
    \EndFunction
    \Function{forget}{$\textbf{Y}, \textbf{C}, \textbf{v}, \textbf{L}$}
      \State $\textbf{v} \gets \textbf{v} - \textbf{Y}_{u}$
      \While {item pair $(i_{1}, i_{2}) \in \textbf{Y}_{u}$}
        \State $C_{i_{1}i_{2}} \gets C_{i_{1}i_{2}}-1$
        \State CPU\_Freq(-1)~~//Tune down DVFS
      \EndWhile
      \While {item $i_{1} \in \textbf{Y}_{u}$}
        \While {item $i_{2} \in \textbf{C}_{i_{1}}$}
          \State $L_{i_{1}i_{2}} \gets C_{i_{1}i_{2}}/(v_{i_{1}}+v_{i_{2}}-C_{i_{1}i_{2}})$
          \State CPU\_Freq(0)~~//Reset DVFS
        \EndWhile
      \EndWhile
    \EndFunction
    \Function{predict}{$j, k, \textbf{L}$}
      \State \Return{top-$k$ items from $\textbf{L}_{j}$}
    \EndFunction
  \end{algorithmic}
\end{algorithm}

\noindent\textbf{Case 1: Personalized PageRank.}
Personalized PageRank (PPR) is a fundamental opration first proposed by Google~\cite{f41}. The PPR algorithm is similar to the recommendation algorithm. They both calculate the distance between users from the perspective of user-item correlation. For example, in web page recommendation, the approach records the browsing history in each device. These pairs of co-occurring pages are ranked and formed the basis for recommendation later.
For the simplest form of PPR, its input includes a binary history matrix $\textbf{Y} \in \left\{0, 1\right\}^{|A|\times|I|}$, which represents the interactions between a set of devices $A$ and a set of items $I$. If the device $j$ interacts with the item $i$, the entry $Y_{ji}$ is equal to 1, and 0 otherwise.

\noindent\emph{Model Construction.} The model of PPR consists of a similarity matrix $\textbf{L} \in {\mathbb{R}}^{|I|\times|I|}$, which represents the interaction similarity between item pairs. A common training method for this model is to first calculate the concurrency matrix $\textbf{C} = \textbf{Y}^{\mathrm{T}} \textbf{Y}$, which represents the number of users interacting with each pair of items. In addition, we need a vector $\textbf{v} = \sum_{j \in U} \textbf{Y}_{j}$ to represent the number of interactions for each item (the sum of the rows of \textbf{Y}). Next, if we want to get the similarity matrix \textbf{L}, we can calculate it by checking the co-occurrence counts. Many similarity measures between items can be computed from the co-occurrence matrix~\cite{f31}. The Jaccard similarity $L_{i_{1}i_{2}}$ between items $i_{1}$ and $i_{2}$ calculated by $L_{i_{1}i_{2}} = C_{i_{1}i_{2}} / (v_{i_{1}}+v_{i_{2}}-C_{i_{1}i_{2}})$ is a better choice. Results can be achieved by querying the similarity matrix $\textbf{L}$, as we described in our motivation example in Figure~\ref{f:moti1}. We retrieve recommendations of item pairs by querying the most similar items in each item, and calculate the preference estimates based on the weighted sum between the similarities of the item and the corresponding user history to generate the items to recommend for a specific device~\cite{f32}.

\noindent\emph{Update Procedure.} We need the following three intermediate data structures including: 1) the item interaction count vector $\textbf{v}$; 2) The concurrency matrix $\textbf{C}$ and 3) the similarity matrix $\textbf{L}$, to enable the incremental and decremental updates for PPR.

We can get the whole process of deleting the u-th user data (corresponding to the $u$-th row $\textbf{Y}_{u}$ in the history matrix $\textbf{Y}$) through the model in the FORGET function (Lines 10-17) of Algorithm 1. We update the corresponding co-occurrence count $C_{i_{1}i_{2}}$ by traversing all item pairs $(i_{1}, i_{2})$ in the user history $\textbf{Y}_{u}$. Finally, we need to 1) traverse each item in the usage history of each user and 2) renew the similarity matrix in corresponding row $\textbf{L}_{i}$. The working principle of the incremental update of the model, explained in the UPDATE function (Lines 2-8) of Algorithm~\ref{alg:1} is similar, the difference is that the number of  simultaneously incremental updates.

\noindent\emph{Space Complexity.} 
PPR is composed of a similarity matrix $\textbf{L} \in {\mathbb{R}}^{|I|\times|I|}$ with the space quadratic of the number of items. As we need to protect the concurrency matrix $\textbf{C} \in {\mathbb{V}}^{|I|\times|I|}$ and the vector $\textbf{v} \in {\mathbb{V}}^{|I|}$, we need to adjust $|\textbf{Y}_{u}|^{2}$ of the concurrency matrix, and the $|\textbf{Y}_{u}|\cdot |I|$ recalculation entries in the similarity matrix \textbf{L} in the worst case. The intermediate data structure of the decremental learning algorithm double the required memory, and the update has a quadratic complexity of $O(|I|^{2})$ in the worst case.  In practice, for example, given a $\theta = 30\%$ configuration and PPR on $I=1000$ items, DEAL uses $\theta$-LRU to  reduce up to 378 page swaps in memory replacement during a single round. However, most users need to interact with very few items in the real world data. In addition, we only retain the top-$k$ entries of each item in \textbf{L}. At the same time, we introduce bounds on the maximum number of interactions to reduce the memory usage required for the intermediate data structures and the update complexity~\cite{f40}. The number of energy control function calls is linear to the number of function calls for incremental/decremental updates. 
Although DEAL   uses the similarity matrix to find corresponding users, it may selectively reduce the data. As the training proceeds, the new data overwrite the old data. Moreover, the new data could   be detected after a few training rounds. The detailed analysis is discussed in Section \ref{SEC:EVA}.

\noindent\emph{Data Recovery.} We analyze how to recover deleted user data from the stale model from the case of deleting the data of a single device from the database. When the original matrix $\textbf{Y}$ removes the row corresponding to the deleted device, the matrix $\textbf{Y}$ becomes an updated matrix $\hat{\textbf{Y}}$. If we still have access to the similarity matrix $\textbf{L}$ calculated from the original matrix $\textbf{Y}$, then we can calculate the corresponding similarity matrix $\hat{\textbf{L}}$ from the updated matrix $\hat{\textbf{Y}}$ and compare it with the stale similarity matrix $\textbf{L}$. All items $i$ with differences in entries of the similarity matrices (e.g., ${\exists}j$ $\textbf{L}_{ij} \neq \hat{\textbf{L}}_{ij}$) are exactly the items that were included in the interaction history of the deleted device. In this way we can recover the deleted data.

\begin{algorithm}[t]
\vspace{0in}
  \caption{Update procedure for Tikhonov Regularization.}
  \label{alg:2}
  \begin{algorithmic}[1]
  \Require $\textbf{z} \gets \textbf{M}^{\mathrm{T}}\textbf{r}, \textbf{QR} \gets qr(\textbf{M}^{\mathrm{T}}\textbf{M}+\lambda\textbf{I})$, user observation $\textbf{M}_{u}$
    \Function{update}{$\textbf{M}_{u}, r_{u}, \textbf{Q}, \textbf{R}, \textbf{z}$}
      \State $ \textbf{z} \gets \textbf{z} + \textbf{M}_{u}r_{u}$
      \State $ \textbf{QR} \gets qr_{update}(\textbf{Q},\textbf{R},\textbf{Q}^{\mathrm{T}}\textbf{M}_{u}, \textbf{M}_{u})$
      \State solve $\textbf{Rh} = \textbf{Q}^{\mathrm{T}}\textbf{z}$ for $\textbf{h}$
      \State CPU\_Freq(1)
    \EndFunction
    \Function{forget}{$\textbf{M}_{u}, r_{u}, \textbf{Q}, \textbf{R}, \textbf{z}$}
      \State $ \textbf{z} \gets \textbf{z} - \textbf{M}_{u}r_{u}$
      \State $ \textbf{QR} \gets qr_{update}(\textbf{Q},\textbf{R},-\textbf{Q}^{\mathrm{T}}\textbf{M}_{u}, \textbf{M}_{u})$
      \State solve $\textbf{Rh} = \textbf{Q}^{\mathrm{T}}\textbf{z}$ for $\textbf{h}$
      \State CPU\_Freq(-1)
    \EndFunction
    \Function{predict}{$\textbf{m}_{new}, \textbf{h}$}
      \State \Return{$\textbf{h}^{\mathrm{T}}\textbf{m}_{new}$}
    \EndFunction
  \end{algorithmic}
\end{algorithm}

\noindent\textbf{Case 2: Tikhonov Regularization.}
Tikhonov regularization~\cite{f33} is a technique widely used to analyze multiple regression data that suffer from multicollinearity. The model input data is the matrix $\textbf{M} \in {\mathbb{R}}^{s \times d}$ of $s$ d-dimensional observations, and the corresponding digital target variable $\textbf{r} \in {\mathbb{R}}^{s}$. 
Under the principle of ensuring generality, we assume that the data of a specific user $i$ is captured in the input $i$-th row vector $\textbf{M}_{i}$. If there is more than one row of data representing a particular user, we can simply perform the decremental update process several times.

\noindent\emph{Model Construction.}
The solution of the normal equation $\textbf{h} = (\textbf{M}^{\mathrm{T}} \textbf{M}+\lambda\textbf{I})^{-1}\textbf{M}^{\mathrm{T}}\textbf{r}$ is a common method to calculate the Tikhonov regularization model in the form of the weight vector $\textbf{h}$. As shown in the PREDICT function (Line 12) of Algorithm \ref{alg:2}, we can use the weight vector as the dot product to calculate the estimate of a new observation $\textbf{m}_{new}$: $\hat{r}_{new} = \textbf{h}^{\mathrm{T}}\textbf{m}_{new}$.

\noindent\emph{Update Procedure.}
We use an effective calculation method to explain the process of deleting the data of a certain device $u$ (corresponding to the $u$-th row $\textbf{M}_{u}$ of matrix $\textbf{M}$) from the Tikhonov regularization model:
\begin{equation}
\textbf{h} = (\textbf{M}^{\mathrm{T}} \textbf{M}-\textbf{M}^{\mathrm{T}}_{u} \textbf{M}+\lambda\textbf{I})^{-1}(\textbf{M}^{\mathrm{T}}\textbf{r}-\textbf{M}_{u}r_{u})
\end{equation}
In this way, we retain two intermediates from the calculation, the vector $\textbf{z} = \textbf{M}^{\mathrm{T}} \textbf{r}$ and a QR factorization $\textbf{QR} = qr(\textbf{M}^{\mathrm{T}} \textbf{M} + \lambda\textbf{I})$ of the regularized gram matrix. We want to use the FORGET function (Lines 7-10) in Algorithm~\ref{alg:2} to solve the updated model $\textbf{h}$. First, we have to use the method of subtracting $\textbf{M}_{u} r_{u}$ to recalculate $\textbf{z}$, and then we have to update the QR decomposition $\textbf{Q}$ and $\textbf{R}$ through using the fast rank-one update algorithm~\cite{f34} with $-\textbf{Q}^{\mathrm{T}} \textbf{M}_{u}$ and $\textbf{M}_{u}$ as parameters. In this way, we can get a new model.

\noindent\emph{Space Complexity.}
We need to maintain two additional matrices $\textbf{Q} \in {\mathbb{R}}^{d \times d}$ and $\textbf{R} \in {\mathbb{R}}^{d \times d}$ as well as the vector $\textbf{z} \in {\mathbb{R}}^{d}$ in the decremental variant of the model. The feature number $d$ of Tikhonov regularization is quadratic and has nothing to do with the number of examples. It is usually much less than the number of examples $m$. 
An decremental/incremental update requires scaling and adding to $\textbf{z}$ ($2d$ operations), the rank-one QR update~\cite{f34} ($26d^{2}$ operations), the matrix vector multiplication $\textbf{Q}^{\mathrm{T}} \textbf{M}_{u}$ ($d^{2}$ operations) and the matrix vector multiplication $\textbf{Q}^{\mathrm{T}} \textbf{z}$ ($d^{2}$ operations), and solving for $\textbf{h}$ by reverse substitution ($d^{2}$ operations). DEAL introduced the FORGET function (to Lines 7-10 of Algorithm \ref{alg:2}). Thus, $\theta$-LRU can significantly reduce more page faults in this linear correlated algorithm. In all, the complexity of our update is $O(d^{2})$, which improves from the original retraining $O(sd^{2})$ complexity.

\noindent\emph{Data Recovery.}
It is  difficult to obtain information about the deleted device feature vector $\textbf{M}_{d}$ from the model $\textbf{h}$.
Although we can constrain the candidate vectors in the subspace defined by $\textbf{h}^{\mathrm{T}} \textbf{M}_{d} = r_{d}$ via accessing the complete target variable $\textbf{r}$, this can produce a large amount of prediction error, so we need to know more about $\textbf{M}$ to further control the candidate vector  space.
\section{evaluation}~\label{SEC:EVA}
In this section, we first introduce the experimental setup and then discuss the corresponding evaluation results from different perspectives in detail. 

\begin{table}[t]
	\caption{Device Profiles} 
	\centering 
	\small
	\begin{tabular}{c c c c} 
		\hline\hline 
		Device & Android Version & $\#$Core & Maximum Frequency \\ [0.5ex] 
		\hline 
		Honor & 8.0 & 8 & 2.11GHz \\ 
		Lenovo & 5.0.2 & 4 & 1.04GHz \\
		ZTE & 5.1.1 & 4 & 1.09GHz \\
		Mi & 5.1.1 & 6 & 1.44GHz \\
		Nexus & 6.0 & 4 & 2.65GHz \\ [1ex] 
		\hline 
	\end{tabular}
	\label{device} 
	\vspace{-0.1in}
\end{table}

\begin{figure*}[t] 
	\centering 
	\begin{subfigure}[t]{0.4\linewidth}
		\includegraphics[width=\linewidth]{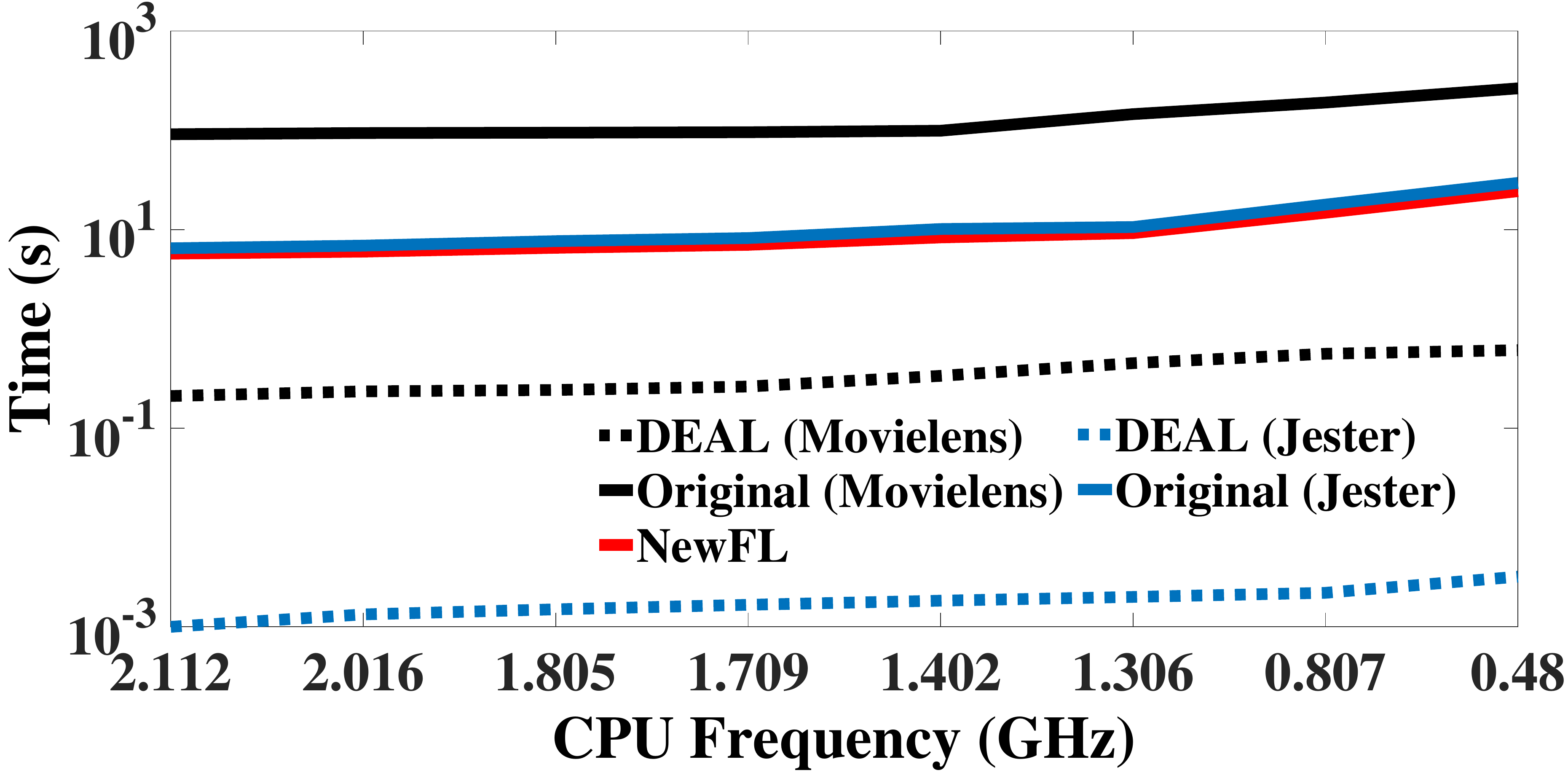}
		\caption{Personalized PageRank.}
		\label{f:time_cf}
	\end{subfigure}
	\begin{subfigure}[t]{0.4\linewidth}
		\includegraphics[width=\linewidth]{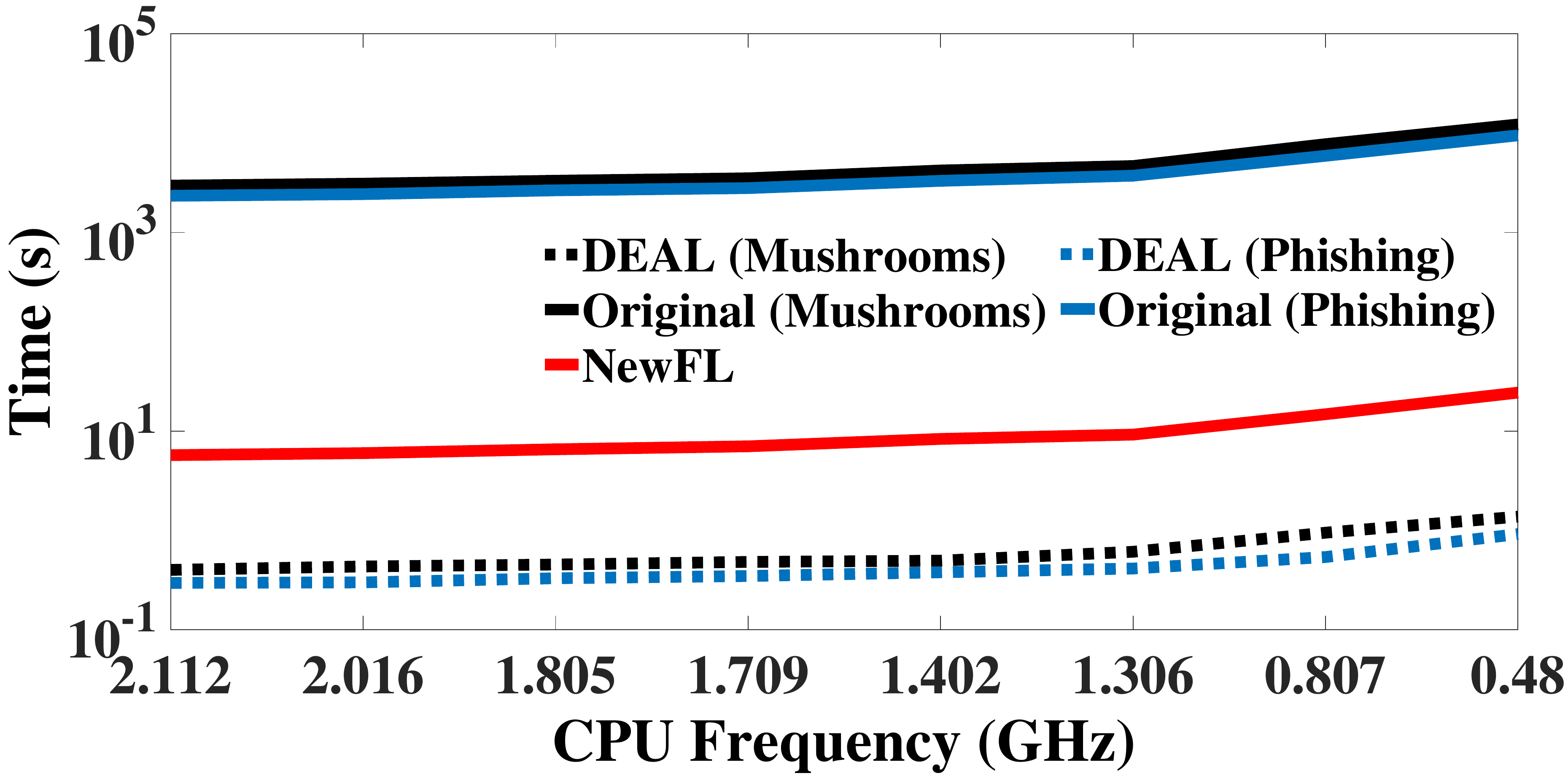}
		\caption{K-Nearest Neighbors.} 
		\label{f:time_lsh}
	\end{subfigure} 
	\begin{subfigure}[t]{0.4\linewidth}
	    \vspace{0.1in}
		\includegraphics[width=\linewidth]{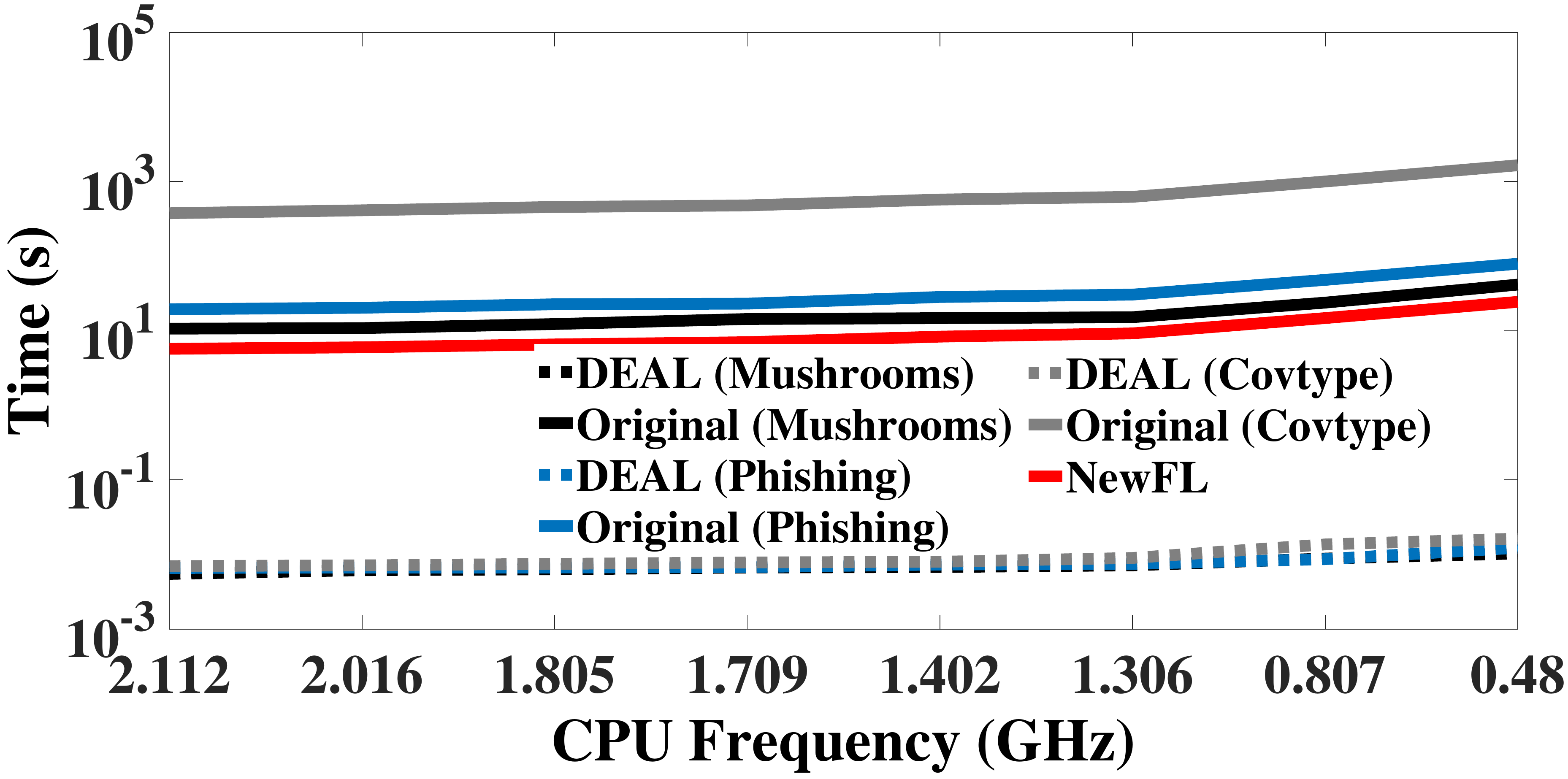}
		\caption{Multinomial Na{\"\i}ve Bayes.}
		\label{f:time_mnb}
	\end{subfigure}
	\begin{subfigure}[t]{0.4\linewidth}
	    \vspace{0.1in}
		\includegraphics[width=\linewidth]{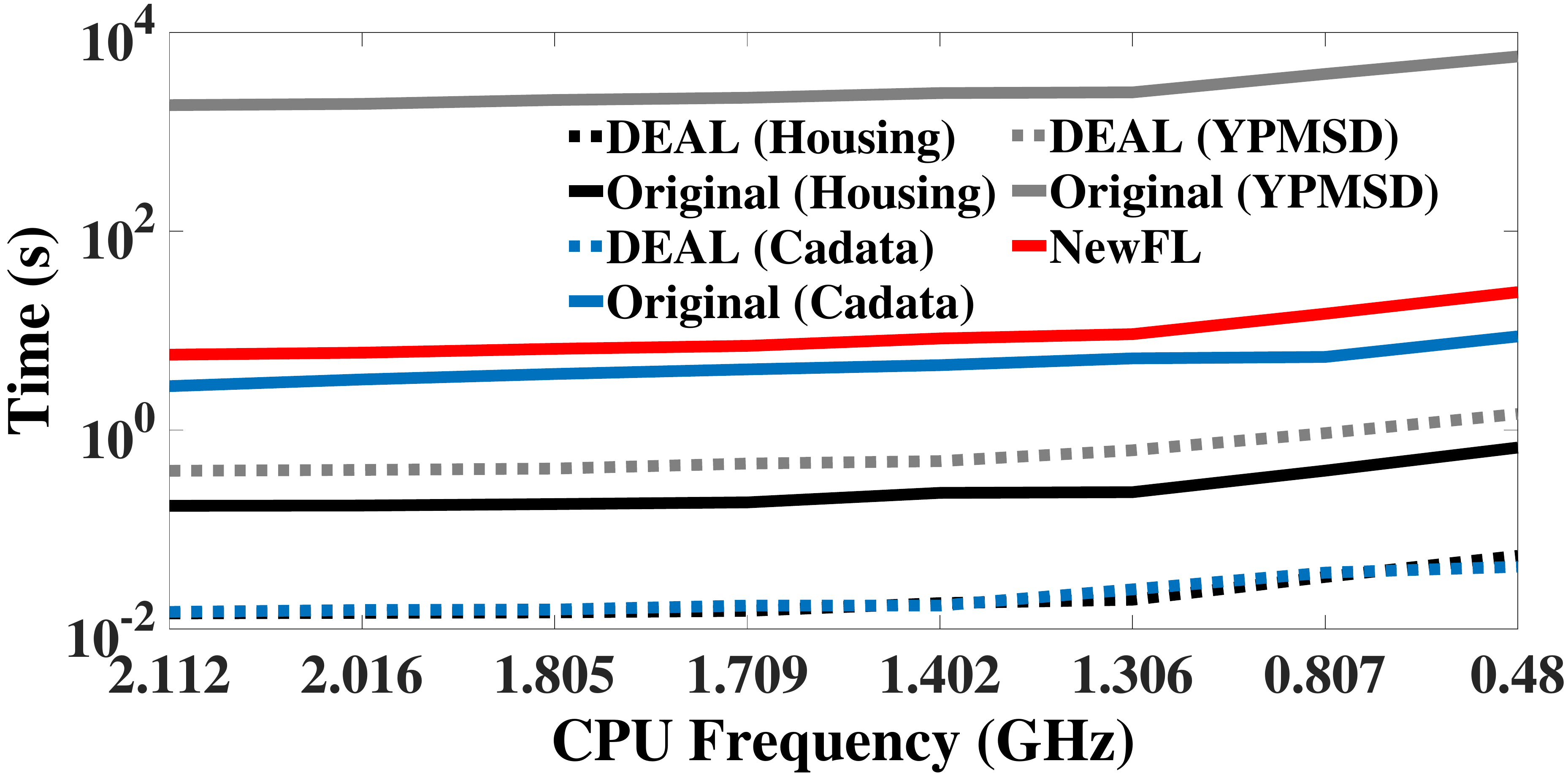}
		\caption{Tikhonov Regularization.} 
		\label{f:time_ridge}
	\end{subfigure} 
	\caption{Training completion time of different application scenarios with different schemes.} 
	\label{f:time} 
	\vspace{-0.2in}
\end{figure*}

\begin{figure}[t]
  \centering
  \includegraphics[width=3in]{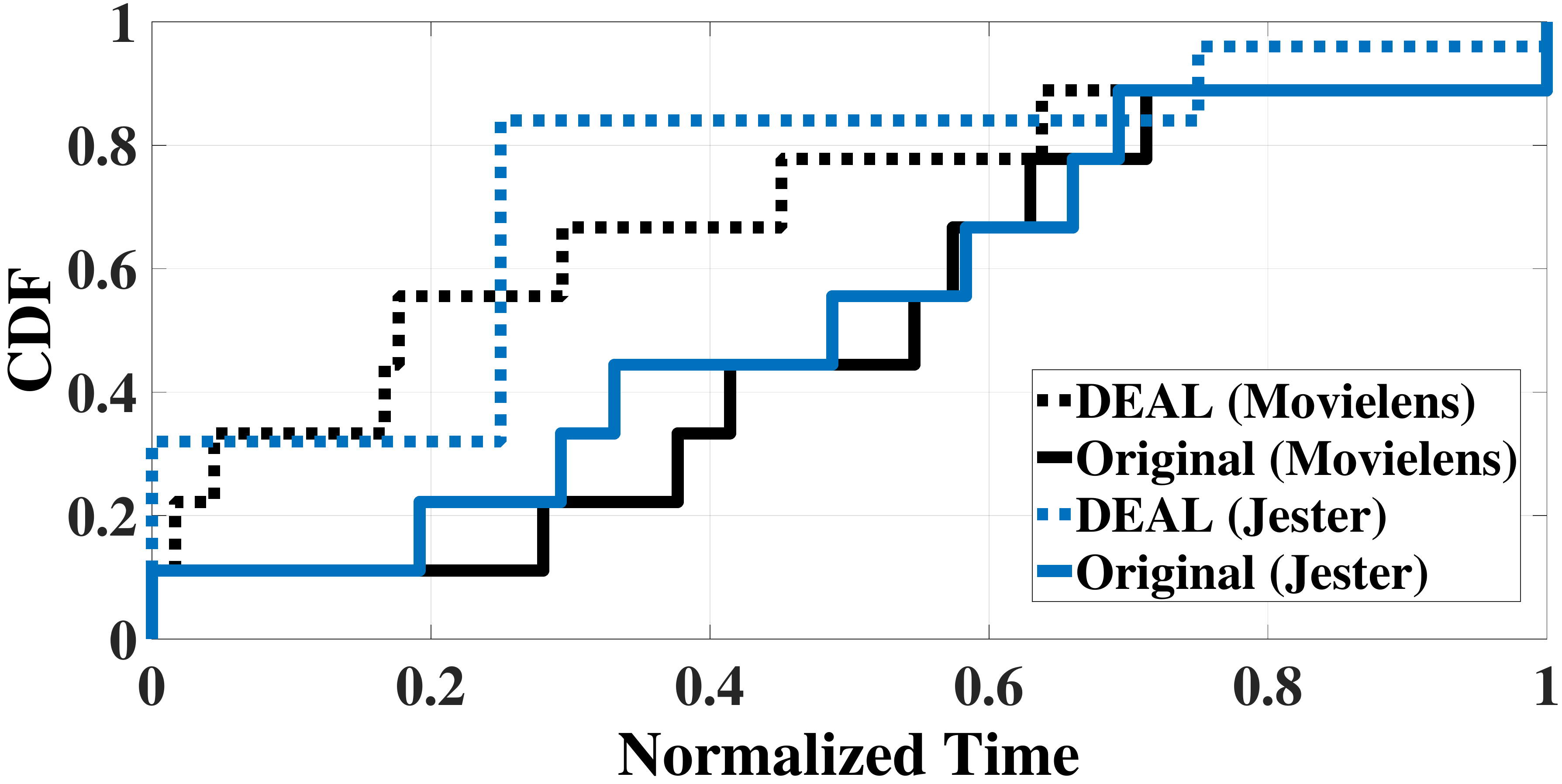}
  \caption{CDF of the convergence time of DEAL and Original in different application scenarios with default governor.}\label{f:cdf}
  \vspace{-0.2in}
\end{figure}

\subsection{Experimental Setup} 
We evaluate the effectiveness of DEAL with both physical testbed and simulation. For the physical testbed, we prototype a federated learning system using mobile devices with different hardware configurations. Table~\ref{device} shows the hardware information of the mobile devices adopted in the experiments. The on-device learning process is implemented based on the deep learning framework DL4J~\cite{f43}. 
In addition, a Monsoon Power Monitor~\cite{f44} is used to measure the power consumption of the participating devices. For the simulation, we emulate different mobile devices with various independent docker images, and deploy hundreds of corresponding FL docker images to simulate corresponding mobile devices. We provide the docker image of the complete experimental code on DockerHub~\footnote{https://hub.docker.com/r/goodlab/deal}. Specifically, in order to evaluate the effectiveness of DEAL, we compare DEAL with the following baselines from different perspectives. 

\begin{itemize}[leftmargin=*]

\item Original, is a federated learning system that always retrain full data objects created for the model.

\item NewFL, is a modified federated learning system implemented using the DL4J framework~\cite{f21}, which only focuses on new data.
\end{itemize}

\noindent\textbf{Models and Datasets:} In order to evaluate the effectiveness of DEAL, we build four different models (Personalized PageRank, K-Nearest Neighbors, Multinomial Na{\"\i}ve Bayes and Tikhonov Regularization) and train them on eight datasets. Specifically, for Personalized PageRank, we use two datasets~\cite{f50} about the movie ratings (movielens) and the joke ratings (jester). For the classification model (K-Nearest Neighbors and Multinomial Na{\"\i}ve Bayes), we use datasets~\cite{f51} about mushrooms, phishing websites (phishing) and cartographic forest data (covtype). For the Tikhonov Regularization model, we used datasets~\cite{f51} on housing prices (housing, cadata) and music (YearPredictionMSD). Finally, in order to understand the impact of training new data only, we adopt image classification dataset Cifar-10. 

\begin{figure}[t]
  \centering
  \includegraphics[width=3in]{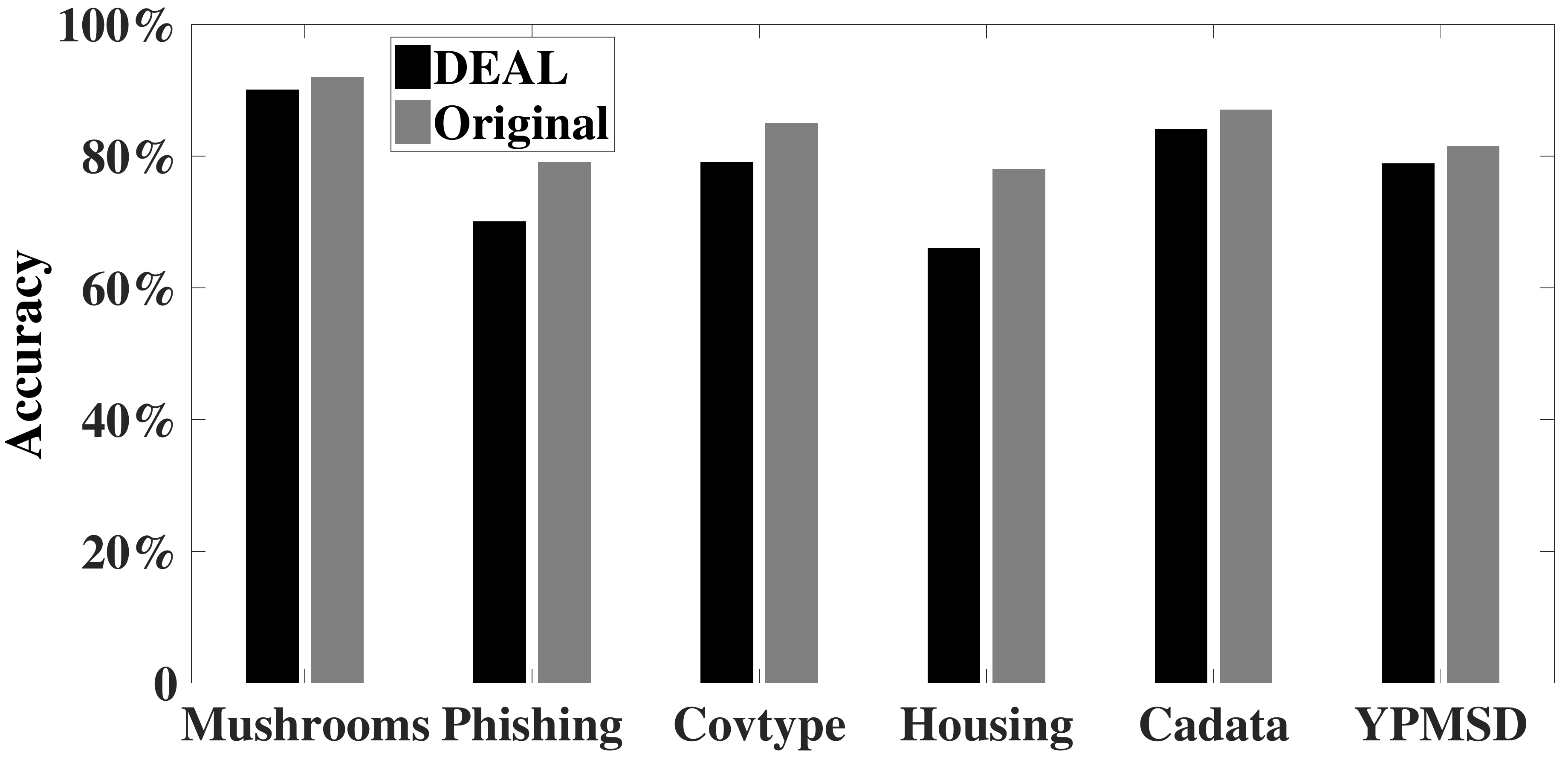}
  \caption{Accuracy of DEAL and Original on Tikhonov Regularization Model with different datasets.}\label{f:accuracy}
  \vspace{-0.2in}
\end{figure}

\begin{figure*}[t] 
	\centering 
	\begin{subfigure}[t]{0.4\linewidth}
		\includegraphics[width=\linewidth]{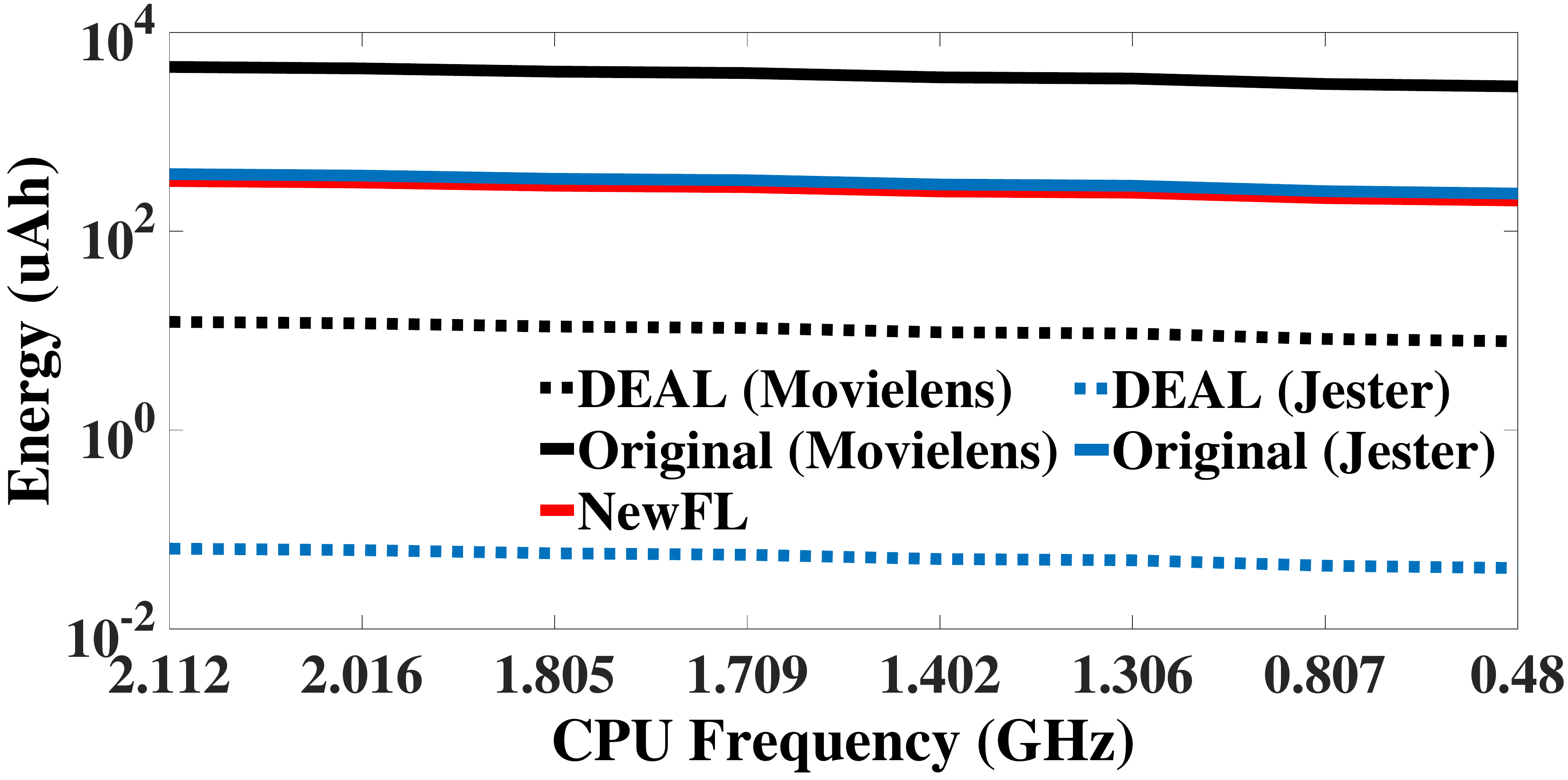}
		\caption{Personalized PageRank.}
		\label{f:energy_cf}
	\end{subfigure}
	\begin{subfigure}[t]{0.4\linewidth}
		\includegraphics[width=\linewidth]{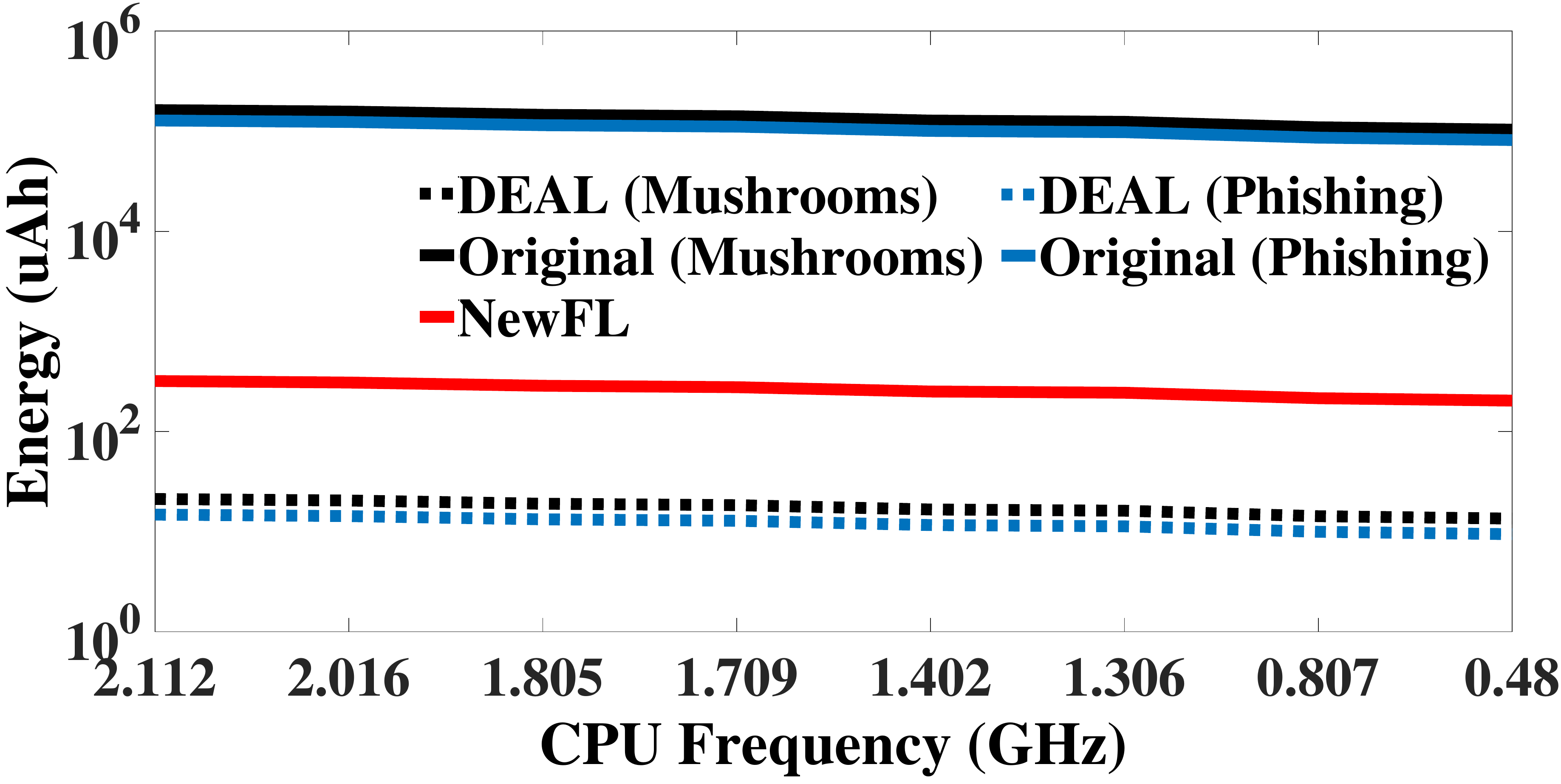}
		\caption{K-Nearest Neighbors.} 
		\label{f:energy_lsh}
	\end{subfigure} 
	\begin{subfigure}[t]{0.4\linewidth}
	    \vspace{0.1in}
		\includegraphics[width=\linewidth]{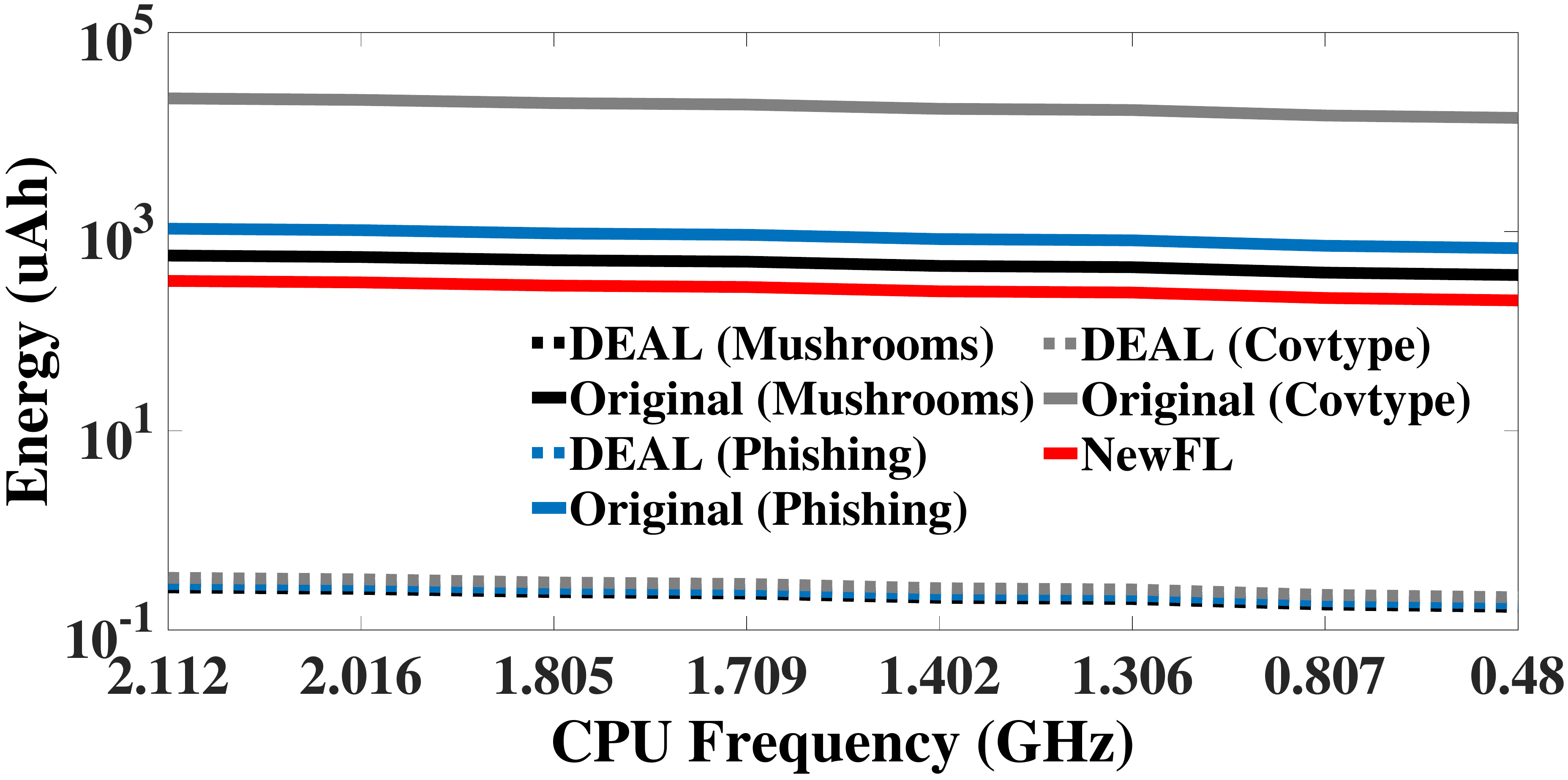}
		\caption{Multinomial Na{\"\i}ve Bayes.}
		\label{f:energy_mnb}
	\end{subfigure}
	\begin{subfigure}[t]{0.4\linewidth}
	    \vspace{0.1in}
		\includegraphics[width=\linewidth]{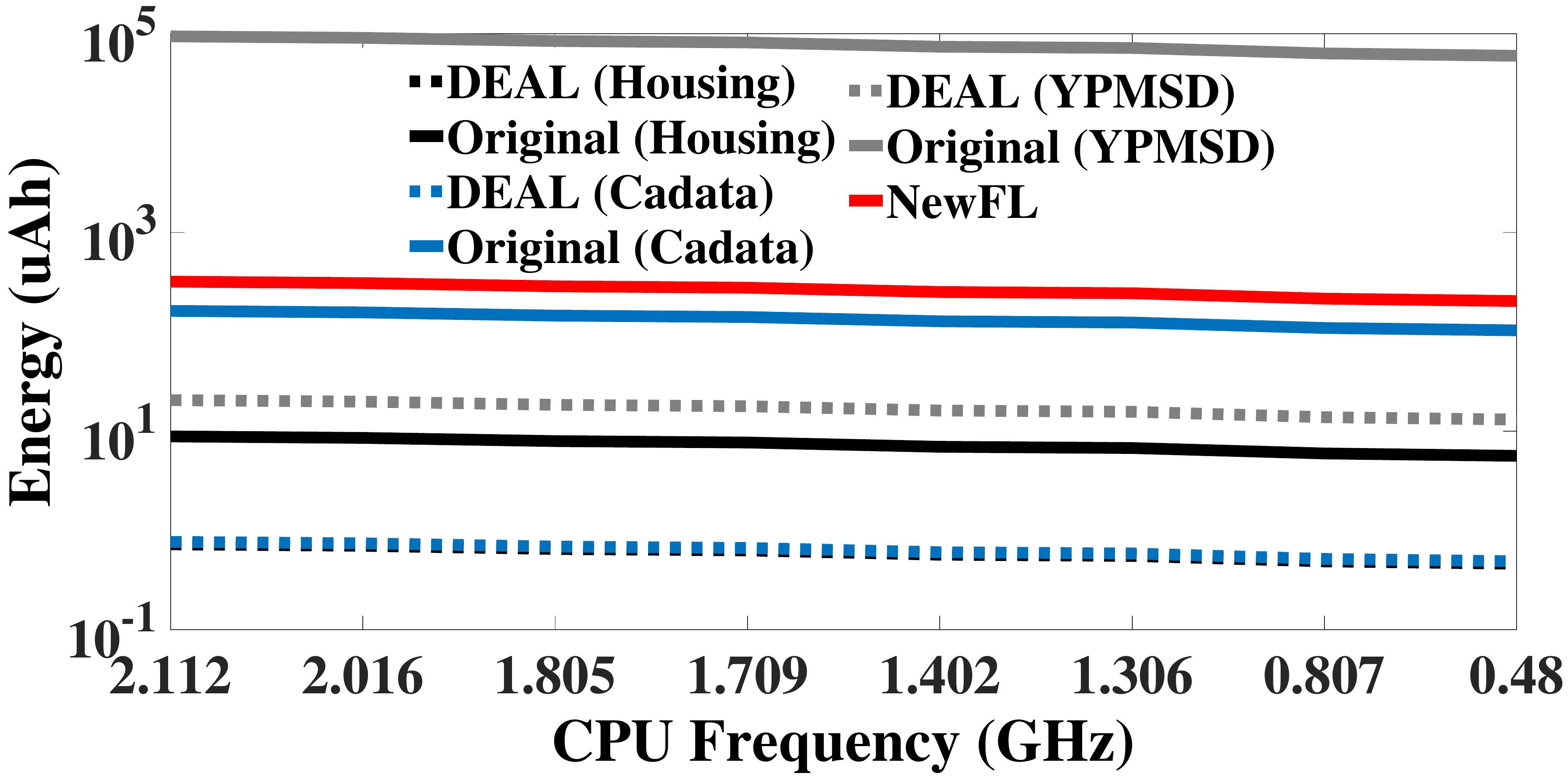}
		\caption{Tikhonov Regularization.} 
		\label{f:energy_ridge}
	\end{subfigure} 
	\caption{Energy consumption during the training process with different schemes in various application scenarios.} 
	\label{f:energy} 
	\vspace{-0.2in}
\end{figure*}

\subsection{Results}
\noindent\textbf{Comparision of Training Completion Time.}
We first train a model on each dataset and load it into the smartphone. Next, we compare the overall training completion time of DEAL with all baselines in different application scenarios, and repeat every experiment for twenty randomly selected users. Figure \ref{f:time} shows the experimental results under different CPU frequencies on Huawei Honor 8 Lite.
Specifically, Figure~\ref{f:time_cf} shows the results of training time of Personalized PageRank model on the movielens and jester datasets. 
In the movielens datasets, DEAL achieves one and two orders of magnitude faster than the NewFL and Original, respectively. 
It is because only DEAL trains less data than the Original, but also DEAL forgets some updates during training, thus greatly reducing the overall training time.

Similarly, Figure~\ref{f:time_lsh} shows the results of training time of a non-linear model, a k-Nearest Neighbor algorithm with Locality Sensitive Hashing, on the mushrooms and phishing datasets. 
DEAL has the similar improvement on the training time, namely one order of magnitude and three orders of magnitude faster than the NewFL and the Original, respectively. 
Moreover, when we allow aggressive DVFS on the device, 
DEAL can achieve four orders of magnitude better performance than the Original, in the phishing dataset. It is because
(1) the phishing dataset is very large and the whole training is more I/O intensive than other workloads, allowing more power saving potentials; (2) The Original always trains the whole data set, which has a much larger memory footprint than DEAL.

In the rest two cases,  Figure~\ref{f:time_mnb} and~\ref{f:time_ridge}, both illustrate that the advantage of DEAL, compared to the two baselines, can achieve 2-4 orders of magnitude faster in training a converged model. However, when the data more coarse-grained distributed such as the YearPredictionMSD dataset,
the performance of DEAL slowly converges to that of the NewFL.
Yet, DEAL always achieves a better modeling performance than the NewFL because DEAL captures feature from older data objects.


\noindent\textbf{Comparision of Convergence Time.}
We deploy hundreds of FL docker images to simulate corresponding mobile devices. Figure~\ref{f:cdf} shows the simulation results. The CDF result of Figure~\ref{f:cdf} shows the trend of the convergence time of DEAL and Original on Personalized PageRank model with the default power governor (interactive). 
Concretely, as can be seen from Figure~\ref{f:cdf}, in the Movielens dataset, 92\% of the simulated devices show that the advantage of DEAL is faster in training the converged model compared with the Original, and the median values of convergence time for DEAL and the Original are 158ms and 94,988ms (normalized to 0.18 and 0.55). In the Jester dataset, 85\% of the simulated devices show that the advantage of DEAL is faster in training the converged model compared with the Original, and the median value of the convergence time for DEAL and the Original are 1ms and 6,598ms (normalized to 0.25 and 0.36). This shows that in our simulation, the convergence time of DEAL of more than half of the mobile devices is three orders of magnitude faster than that of the Original.
This is because DEAL allows the server to communicate with workers via the SUB method periodically, and start the central convergence when receiving more than half SUB signals from all selected workers or passed a TTL. DEAL does not need to wait for all models update. However, it can be seen that the effect of the Tail of the convergence time is not good enough.

\noindent\textbf{Comparison of Model Accuracy.}
Figure~\ref{f:accuracy} compares the accuracy of the Tikhonov regularization model on different datasets. It shows that in the phishing dataset, the model accuracy of DEAL is only 9\% lower than that of the Original. Among these datasets, the housing dataset has the largest accuracy reduction, which is reduced by 12\%. The accuracy of the remaining datasets is not much different from the Original, which are all about 3\%.
The results show that, while guaranteeing the accuracy of the model, DEAL effectively improves the learning speed for on-device federated learning, and it can be seen from the subsequent results that DEAL also improves the energy efficiency of federated learning on the devices.

\noindent\textbf{Comparison of Energy Consumption.}
Rather than having a less training time, DEAL allows adaptive power control in the training algorithm, as aforementioned in Section~\ref{SEC:DES}. Here, we provide an energy consumption analysis based on our measured data under different CPU frequencies on the Huawei Honor 8 Lite in Figure~\ref{f:energy}.
One common theme behind the power saving is, no matter which baselines, the total energy consumed  gradually decreases with the CPU frequency.

Figure~\ref{f:energy_cf} shows the results of the consumed energy for training of Personalized PageRank model on the movielens and jester datasets. It can be seen that DEAL of the movielens dataset saves 253.2uAh of energy and 3687.1uAh of energy, as compared to the NewFL and the original, respectively. DEAL can save about 300uAh in the jester dataset. Figure~\ref{f:energy_lsh} shows the results of the consumed energy for training of k-Nearest Neighbors with Locality Sensitive Hashing  on the mushrooms and phishing datasets. It can be seen that DEAL of the mushrooms and phishing datasets consume an order of magnitude less energy than the NewFL, saving about 250uAh of energy. Compared to the Original, DEAL achieves energy saving in the amount of approximately 110,000uAh. Figure~\ref{f:energy_mnb} shows the results of the consumed energy for training  Multinomial Na{\"\i}ve Bayes models on mushrooms, phishing, and covtype datasets. DEAL of these three datasets consumes two to three orders of magnitude less energy than the NewFL, saving about 263uAh of energy. DEAL of the mushrooms and phishing datasets consumes three orders of magnitude less energy than the Original. In the the covtype dataset, DEAL consumes 4 orders of magnitude less energy than the Original. Because the cardinality of the covtype dataset is much larger than the mushrooms and phishing datasets, the training time and power consumption required for a whole retraining increase accordingly. Specifically, DEAL of the covtype datasets saves 17,908.1uAh of energy compared to the Original. Figure~\ref{f:energy_ridge} shows the results of the consumed energy for training of Tikhonov regularization model on the housing, cadata, and YearPredictionMSD datasets. DEAL of the housing dataset saves only 6.7uAh of energy compared to the Original. This is because the housing dataset size is too small, so it consumes less energy for retraining. DEAL of the YearPredictionMSD dataset saves 77,497.6uAh of energy compared to the Original. 

\begin{figure}[t]
  \vspace{-0.01in}
  \centering
  \includegraphics[width=3in]{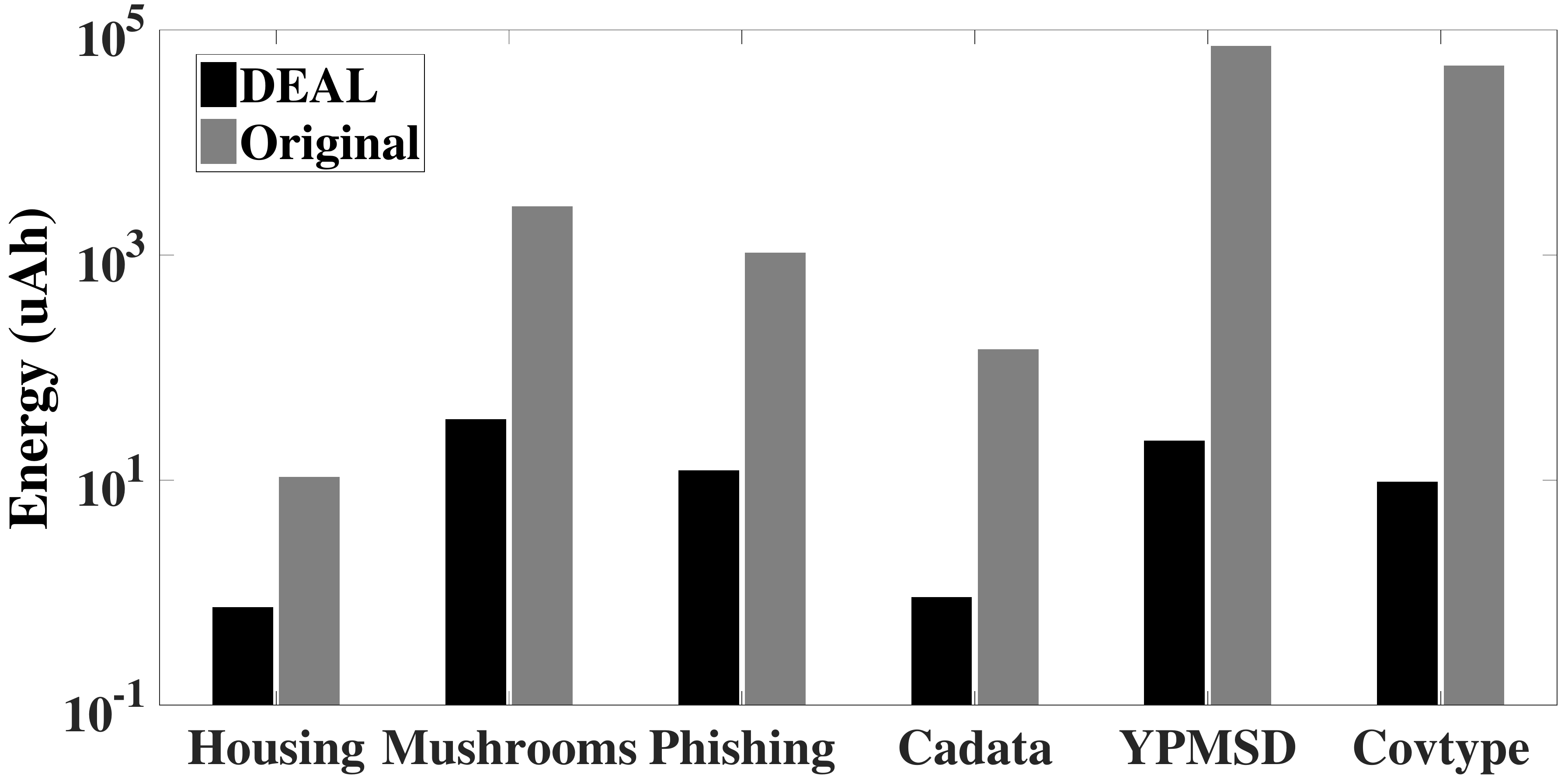}
  \caption{Energy consumption of DEAL and Original on Tikhonov Regularization Model with different datasets.}\label{f:energyanddatasets}
  \vspace{-0.2in}
\end{figure}

Figure~\ref{f:energyanddatasets} compares the energy consumption of DEAL and Original on the Tikhonov regularization model for six different datasets: housing, mushrooms, phishing, cadata, YearPreditionMSD and covtype. It can be seen that no matter what kind of dataset, DEAL consumes more than one order of magnitude less energy compared to the Original. Some datasets can even save three orders of magnitude of energy.

In short, DEAL can save up to 81.7\% and 80.6\% of energy cost on average, compared to the Original and the NewFL, respectively. With the smallest dataset housing in the Tikhonov regularization model, DEAL still saves 75.6\% of energy, as compared to the Original. And due to the different size of datasets in each model, the behave of DEAL can change accordingly when using different datasets.

\begin{figure}[t]
  \vspace{-0.01in}
  \centering
  \includegraphics[width=3in]{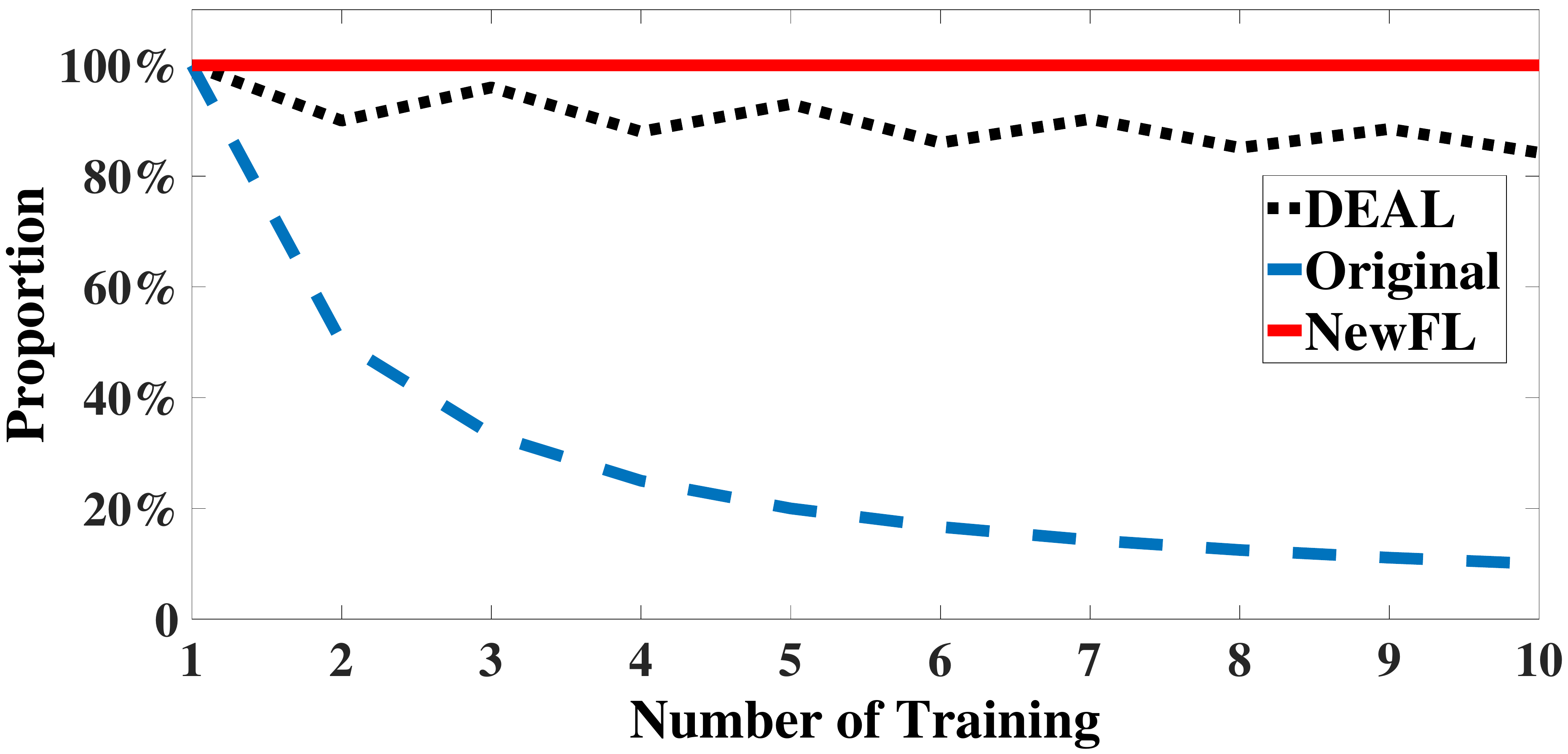}
  \caption{Privacy under three different baselines.}\label{f:privacy}
  \vspace{-0.2in}
\end{figure}

\noindent\textbf{Comparison of Privacy.}
There is currently no existing approach that can effectively quantify privacy on mobile phones ~\cite{f47,f48}, so we measure privacy by observing the proportion of data objects. We add 10 new data objects in each round of training and observe the proportion of these 10 new data objects to the overall training data objects to measure the privacy situation. 
It can be seen from Figure~\ref{f:privacy} that for NewFL, its effect is the best, it only trains new data, so its proportion is always 100\%. For Original, because it needs to train all the data (10 newly added data and the previous old data), as the number of training increases, its proportion value continues to decrease. For DEAL, there is a phenomenon of jitter, because DEAL includes two learning methods: decremental learning and incremental learning.
In real life, we generally delete old data, so DEAL pays less and less attention to old data. In addition, new data always overwrites old data. But when we need to delete new data, DEAL can also delete these data in a specific training round.

\section{Related Work}~\label{SEC:relatedwork}
Our work is closely related to two major research topics, \textit{distributed learning} and \textit{federated learning}. 

\noindent\textbf{Distributed Learning.} Distributed learning has attracted a lot of attention in order to effectively train different neural network models with large amount of data located at different places. Previous research has been done to improve the system performance of distributed learning from different perspectives. Zhang et al. \cite{f10} design a scheduling algorithm to approximate  the training performance of deep learning jobs in order to maximize the overall performance of tasks in a cluster. Li et al. \cite{f12} propose a framework of parameter server for distributed learning in order to manage asynchronous data communication between different working nodes and support flexible consistency models, elastic scalability and continuous fault tolerance. 
Though these approaches can effectively improve the system performance of distributed learning, they cannot be directly applied to mobile based federated learning. Compared with servers located in the data center, mobile devices have much higher limitation of computing capacity and battery lifetime.

\noindent\textbf{Federated Learning.} Federated learning is proposed to make multiple mobile devices collaboratively train a shared deep learning model while guaranteeing the data privacy \cite{f1,f2,f3,f4,f5,f6,f7,f8,f9,f36}. Lalitha et al. \cite{f1} design a distributed learning algorithm to train a machine learning model over a network of users in a fully decentralized manner. 
Bonawitz et al. \cite{f9} build a salable production system for federated learning in the domain of mobile devices, based on Tensorflow.
Konecny et al. \cite{f2} aim to improving the communication efficiency in a federated learning system and propose two schemes (e.g., sketched update and structured update) to reduce the uplink communication costs. Smith et al. \cite{f3} propose a system-aware optimization approach to solve problems of high communication cost, stragglers, and fault tolerance for distributed multi-task learning. Wu et al. \cite{f36} adopt a data-driven approach to introduce the opportunities and design challenges faced by Facebook in order to enable machine learning inference locally on smartphones and other edge platforms.
However, the problem of effectively reducing the energy consumption while guaranteeing the model accuracy is not sufficiently investigated which is critical to battery-powered mobile device. 

\section{Conclusion}~\label{SEC:CON}
This paper proposes an energy efficient learning framework, DEAL, that achieves energy saving with a decremental learning design. DEAL improves the energy efficiency of the training process from two main levels. The first level selects a subset of workers with sufficient capacity in order to maximize the rewards, i.e., energy saving potentials. 
The second level is made up of a specified decremental learning algorithm that actively provides a decremental and incremental update functions, which adaptively tunes the DVFS of local mobile device. DEAL is prototyped in containerized services with modern smartphone profiles and evaluated with different learning benchmarks with real-world traces.
The evaluation result shows that DEAL achieves 75.6\%--82.4\% less energy footprint in different datasets, compared to the traditional methods. Moreover, all learning processes are faster than the classic federated learning framework up to 2-4 orders of magnitude. Immediate future work includes evaluating DEAL with 
more applications and models on more smartphones at scaling-out.  

{\footnotesize
\bibliographystyle{IEEEtran}
\bibliography{biblio}

\begin{thebibliography}{10}
\providecommand{\url}[1]{#1}
\csname url@samestyle\endcsname
\providecommand{\newblock}{\relax}
\providecommand{\bibinfo}[2]{#2}
\providecommand{\BIBentrySTDinterwordspacing}{\spaceskip=0pt\relax}
\providecommand{\BIBentryALTinterwordstretchfactor}{4}
\providecommand{\BIBentryALTinterwordspacing}{\spaceskip=\fontdimen2\font plus
\BIBentryALTinterwordstretchfactor\fontdimen3\font minus
  \fontdimen4\font\relax}
\providecommand{\BIBforeignlanguage}[2]{{%
\expandafter\ifx\csname l@#1\endcsname\relax
\typeout{** WARNING: IEEEtran.bst: No hyphenation pattern has been}%
\typeout{** loaded for the language `#1'. Using the pattern for}%
\typeout{** the default language instead.}%
\else
\language=\csname l@#1\endcsname
\fi
#2}}
\providecommand{\BIBdecl}{\relax}
\BIBdecl

\bibitem{f1}
A.~Lalitha \emph{et~al.}, ``Fully decentralized federated learning,'' in
  \emph{Third workshop on Bayesian Deep Learning (NeurIPS)}, 2018.

\bibitem{f22}
G.~C. Burdea and P.~Coiffet, \emph{Virtual reality technology}.\hskip 1em plus
  0.5em minus 0.4em\relax John Wiley \& Sons, 2003.

\bibitem{f23}
S.~Khattar \emph{et~al.}, ``Smart home with virtual assistant using raspberry
  pi,'' in \emph{2019 9th International Conference on Cloud Computing, Data
  Science \& Engineering (Confluence)}.\hskip 1em plus 0.5em minus 0.4em\relax
  IEEE, 2019, pp. 576--579.

\bibitem{f24}
B.~Nemade, ``Automatic traffic surveillance using video tracking,''
  \emph{Procedia Computer Science}, vol.~79, pp. 402--409, 2016.

\bibitem{f25}
``Big data market worth \$229.4 billion by 2025,''
  \url{https://www.marketsandmarkets.com/PressReleases/big-data.asp}.

\bibitem{f48}
T.~Li \emph{et~al.}, ``Federated learning: Challenges, methods, and future
  directions,'' \emph{IEEE Signal Processing Magazine}, vol.~37, no.~3, pp.
  50--60, 2020.

\bibitem{f26}
B.~McMahan and D.~Ramage, ``Federated learning: Collaborative machine learning
  without centralized training data,'' \emph{Google Research Blog}, vol.~3,
  2017.

\bibitem{f27}
J.~Zhou \emph{et~al.}, ``Capman: Cooling and active power management in big.
  little battery supported devices,'' EasyChair, Tech. Rep., 2020.

\bibitem{f28}
``Dataset,'' \url{https://data.world/crowdflower/ecommerce-search-relevance}.

\bibitem{f29}
``Gdpr.eu. recital 65: Right of rectification and erasure,''
  \url{https://gdpr.eu/recital-65-right-of-rectification-and-erasure}.

\bibitem{f42}
S.~Bag \emph{et~al.}, ``An efficient recommendation generation using relevant
  jaccard similarity,'' \emph{Information Sciences}, vol. 483, pp. 53--64,
  2019.

\bibitem{f30}
Z.~Xu \emph{et~al.}, ``Exploring federated learning on battery-powered
  devices,'' in \emph{Proceedings of the ACM Turing Celebration
  Conference-China}, 2019, pp. 1--6.

\bibitem{f9}
K.~Bonawitz \emph{et~al.}, ``Towards federated learning at scale: System
  design,'' \emph{arXiv preprint arXiv:1902.01046}, 2019.

\bibitem{f40}
S.~Schelter, ``“amnesia”-a selection of machine learning models that can
  forget user data very fast,'' \emph{suicide}, vol. 8364, no. 44035, p. 46992.

\bibitem{f20}
G.~Cauwenberghs and T.~Poggio, ``Incremental and decremental support vector
  machine learning,'' \emph{Advances in neural information processing systems},
  vol.~13, pp. 409--415, 2000.

\bibitem{f39}
A.~Pathak \emph{et~al.}, ``Where is the energy spent inside my app? fine
  grained energy accounting on smartphones with eprof,'' in \emph{Proceedings
  of the 7th ACM european conference on Computer Systems}, 2012, pp. 29--42.

\bibitem{f45}
``Prior work,'' \url{https://github.com/good-ncu/Appendix}.

\bibitem{f46}
F.~Li \emph{et~al.}, ``Combinatorial sleeping bandits with fairness
  constraints,'' \emph{IEEE Transactions on Network Science and Engineering},
  2019.

\bibitem{f49}
J.~Gittins \emph{et~al.}, \emph{Multi-armed bandit allocation indices}.\hskip
  1em plus 0.5em minus 0.4em\relax John Wiley \& Sons, 2011.

\bibitem{f35}
Y.~Hou \emph{et~al.}, ``Course recommendation of mooc with big data support: A
  contextual online learning approach,'' in \emph{IEEE INFOCOM 2018-IEEE
  Conference on Computer Communications Workshops (INFOCOM WKSHPS)}.\hskip 1em
  plus 0.5em minus 0.4em\relax IEEE, 2018, pp. 106--111.

\bibitem{f41}
S.~Wang \emph{et~al.}, ``Fora: simple and effective approximate single-source
  personalized pagerank,'' in \emph{Proceedings of the 23rd ACM SIGKDD
  International Conference on Knowledge Discovery and Data Mining}, 2017, pp.
  505--514.

\bibitem{f31}
S.~Schelter \emph{et~al.}, ``Scalable similarity-based neighborhood methods
  with mapreduce,'' in \emph{Proceedings of the sixth ACM conference on
  Recommender systems}, 2012, pp. 163--170.

\bibitem{f32}
B.~Sarwar \emph{et~al.}, ``Item-based collaborative filtering recommendation
  algorithms,'' in \emph{Proceedings of the 10th international conference on
  World Wide Web}, 2001, pp. 285--295.

\bibitem{f33}
C.~Groetsch, ``The theory of tikhonov regularization for fredholm equations,''
  \emph{104p, Boston Pitman Publication}, 1984.

\bibitem{f34}
G.~H. Golub and C.~F. Van~Loan, \emph{Matrix computations}.\hskip 1em plus
  0.5em minus 0.4em\relax JHU press, 2012, vol.~3.

\bibitem{f43}
A.~C. Nicholson and A.~Gibson, ``Deeplearning4j: Open-source, distributed deep
  learning for the jvm,'' \emph{Deeplearning4j. org}, 2017.

\bibitem{f44}
``Monsoon power monitor,''
  \url{http://www.msoon.com/LabEquipment/PowerMonitor/}.

\bibitem{f21}
C.~J, ``Federated learning - proandroiddev,''
  \url{https://proandroiddev.com/federated-learning-e79e054c33ef}.

\bibitem{f50}
``Personalized pagerank datasets,'' \url{http://konect.cc/networks/}.

\bibitem{f51}
``Classification and regression datasets,''
  \url{https://www.csie.ntu.edu.tw/~cjlin/libsvmtools/datasets/}.

\bibitem{f47}
V.~Mothukuri \emph{et~al.}, ``A survey on security and privacy of federated
  learning,'' \emph{Future Generation Computer Systems}, vol. 115, pp.
  619--640, 2020.

\bibitem{f10}
H.~Zhang \emph{et~al.}, ``Slaq: quality-driven scheduling for distributed
  machine learning,'' in \emph{Proceedings of the 2017 Symposium on Cloud
  Computing}, 2017, pp. 390--404.

\bibitem{f12}
M.~Li \emph{et~al.}, ``Scaling distributed machine learning with the parameter
  server,'' in \emph{11th $\{$USENIX$\}$ Symposium on Operating Systems Design
  and Implementation ($\{$OSDI$\}$ 14)}, 2014, pp. 583--598.

\bibitem{f2}
J.~Kone{\v{c}}n{\`y} \emph{et~al.}, ``Federated learning: Strategies for
  improving communication efficiency,'' \emph{arXiv preprint arXiv:1610.05492},
  2016.

\bibitem{f3}
V.~Smith \emph{et~al.}, ``Federated multi-task learning,'' in \emph{Advances in
  Neural Information Processing Systems}, 2017, pp. 4424--4434.

\bibitem{f4}
B.~McMahan \emph{et~al.}, ``Communication-efficient learning of deep networks
  from decentralized data,'' in \emph{Artificial Intelligence and
  Statistics}.\hskip 1em plus 0.5em minus 0.4em\relax PMLR, 2017, pp.
  1273--1282.

\bibitem{f5}
K.~Bonawitz \emph{et~al.}, ``Practical secure aggregation for
  privacy-preserving machine learning,'' in \emph{Proceedings of the 2017 ACM
  SIGSAC Conference on Computer and Communications Security}, 2017, pp.
  1175--1191.

\bibitem{f6}
M.~R. Sprague \emph{et~al.}, ``Asynchronous federated learning for geospatial
  applications,'' in \emph{Joint European Conference on Machine Learning and
  Knowledge Discovery in Databases}.\hskip 1em plus 0.5em minus 0.4em\relax
  Springer, 2018, pp. 21--28.

\bibitem{f7}
F.~Mo and H.~Haddadi, ``Efficient and private federated learning using tee,''
  in \emph{EuroSys}, 2019.

\bibitem{f8}
S.~Wang \emph{et~al.}, ``Adaptive federated learning in resource constrained
  edge computing systems,'' \emph{IEEE Journal on Selected Areas in
  Communications}, vol.~37, no.~6, pp. 1205--1221, 2019.

\bibitem{f36}
C.-J. Wu \emph{et~al.}, ``Machine learning at facebook: Understanding inference
  at the edge,'' in \emph{2019 IEEE International Symposium on High Performance
  Computer Architecture (HPCA)}.\hskip 1em plus 0.5em minus 0.4em\relax IEEE,
  2019, pp. 331--344.

\end{thebibliography}
}

\end{document}